\documentclass{article}

\usepackage{microtype}
\usepackage{graphicx}

\usepackage{booktabs}

\usepackage{hyperref}

\usepackage[accepted]{icml2025}

\usepackage{amsmath}
\usepackage{amssymb}
\usepackage{mathtools}
\usepackage{amsthm}

\usepackage[capitalize,noabbrev]{cleveref}

\usepackage{xcolor}

\usepackage{subcaption}

\renewcommand{\eqref}[1]{Eq.~(\ref{eq:#1})}
\newcommand{\figref}[1]{Figure \ref{fig:#1}}
\newcommand{\tabref}[1]{Table \ref{tab:#1}}        
\newcommand{\secref}[1]{Section \ref{sec:#1}}
\newcommand{\appref}[1]{Appendix \ref{app:#1}}
\newcommand{\myalgref}[1]{Alg.~\ref{alg:#1}}
\renewcommand{\P}{\mathbb{P}}
\newcommand{\reals}{\mathbb{R}}
\newcommand{\one}{\mathbb{I}}
\newcommand{\norm}[1]{\|#1\|}
\newcommand{\st}{\text{ s.t. }}

\newcommand{\cA}{\mathcal{A}}

\newcommand{\cC}{\mathcal{C}}
\newcommand{\cD}{\mathcal{D}}

\newcommand{\cN}{\mathcal{N}}

\newcommand{\cP}{\mathcal{P}}

\newcommand{\cY}{\mathcal{Y}}

\newcommand{\ta}{\alpha_a}

\DeclareMathOperator*{\Minimize}{Minimize}

\newcommand{\hp}{\hat{p}}
\newcommand{\bhp}{\hat{\mathbf{p}}}
\newcommand{\bpi}{{\boldsymbol\pi}}

\newcommand{\base}{{\textsf{b}}}
\newcommand{\alla}{\boldsymbol{\alpha}_\base}
\newcommand{\allag}{\boldsymbol{\alpha}_a}
\newcommand{\allabigg}{\boldsymbol{\alpha}_{\cA}}
\newcommand{\matdelta}{\Delta_{k\times k}}

\newcommand{\minunf}{\texttt{min\UNF}}

\newcommand{\dbase}{\cD_{\base}}
\newcommand{\alphabase}{\alpha_{\base}}
\newcommand{\matalpha}{\mathbf{M}}

\newcommand{\pos}{\pi}
\newcommand{\confg}{\matalpha_a}
\newcommand{\confgall}{\matalpha_{\cA}}

\newcommand{\bfalpha}{\boldsymbol{\alpha}}

\newcommand{\bfx}{\mathbf{x}}

\newcommand{\bfc}{\mathbf{c}}

\newcommand{\bfv}{\mathbf{v}}

\newcommand{\bfe}{\mathbf{e}}

\newcommand{\bfH}{\mathbf{H}}

\newcommand{\bw}{\mathbf{w}}
\newcommand{\bv}{\mathbf{v}}
\newcommand{\bfw}{\mathbf{\widetilde{w}}}
\newcommand{\bfJ}{{\mathbf J}}

\newcommand{\bfM}{{\mathbf M}}

\newcommand{\bfeta}{{\boldsymbol \eta}}

\newcommand{\tw}{\widetilde{w}}
\newcommand{\bof}{\mathbf{f}}

\newcommand{\githublink}{\url{https://github.com/sivansabato/DCPmulticlass}}

\theoremstyle{plain}
\newtheorem{theorem}{Theorem}[section]

\theoremstyle{definition}

\theoremstyle{remark}

\usepackage[textsize=tiny]{todonotes}

\newcommand{\unfairnessname}{Disparate Conditional Prediction}
\newcommand{\UNF}{\ensuremath{\mathrm{DCP}}}

\icmltitlerunning{Disparate Conditional Prediction in Multiclass Classifiers}

\begin{document}

\twocolumn[
\icmltitle{Disparate Conditional Prediction in Multiclass Classifiers}

\begin{icmlauthorlist}
\icmlauthor{Sivan Sabato}{mcmaster,bgu}
\icmlauthor{Eran Treister}{bgu}
\icmlauthor{Elad Yom-Tov}{biu}
\end{icmlauthorlist}

\icmlaffiliation{mcmaster}{Department of Computing and Software, McMaster University; Canada CIFAR AI Chair, Vector Institute}
\icmlaffiliation{biu}{Department of Computer Science, Bar-Ilan University}
\icmlaffiliation{bgu}{Department of Computer Science, Ben-Gurion University of the Negev}

\icmlcorrespondingauthor{Sivan Sabato}{sabatos@mcmaster.ca}

\icmlkeywords{fairness, classification, multiclass, disparate conditional prediction}

\vskip 0.3in
]

\printAffiliationsAndNotice{} 

\begin{abstract}
  We propose methods for auditing multiclass classifiers for fairness under multiclass equalized odds, by estimating the deviation from equalized odds when the classifier is not completely fair. We generalize to multiclass classifiers the measure of \unfairnessname~(\UNF), originally suggested by \citet{SabatoYo20} for binary classifiers. \UNF~is defined as the fraction of the population for which the classifier predicts with conditional prediction probabilities that differ from the closest common baseline. We provide new local-optimization methods for estimating the multiclass \UNF under two different regimes, one in which the conditional confusion matrices for each protected sub-population are known, and one in which these cannot be estimated, for instance, because the classifier is inaccessible or because good-quality individual-level data is not available. These methods can be used to detect classifiers that likely treat a significant fraction of the population unfairly. Experiments demonstrate the accuracy of the methods. Code is provided at \githublink.

\end{abstract}

\section{Introduction}\label{sec:intro}

Fairness of classifiers is a crucial property in many real-life scenarios \citep[see, e.g.,][]{caton2024fairness}. In particular, \emph{auditing} classifiers for fairness is essential in a wide range of applications and has been studied in many works \citep[e.g.,][]{saleiro2018aequitas,angwin2022machine,taskesen2021statistical,cherian2024statistical}. 

While it is desirable that a classifier accurately satisfies the required fairness criterion, this is in many cases not achievable, due to inherent limitations of the distribution \citep{PleissRaKlKi17, KleinbergMuRa17, MenonWi18, wang2024aleatoric},  as well as practical constraints. It is thus necessary to be able to quantify the \emph{deviation} from fairness of a given classifier.
In addition, often the classifier is proprietary \citep{dastin2022amazon,grandinetti2023examining} and thus is not directly accessible for auditing. In other cases, individual-level data that is required for auditing is missing or insufficient. For instance, a health insurance company may use a proprietary classifier to decide on coverage or on premium rates \citep{kafuria2022predictice}. \citet{SabatoYo20} showed that for binary classifiers, it is possible to provide fairness auditing without individual-level data, using only population-level information on the frequencies of positive predictions and of true positive labels in each sub-population, where a sub-population is the group of individuals who have the same value of the protected attribute(s). They considered fairness in the sense of \emph{equalized odds} \citep{HardtPrSr16}, which defines a binary classifier as fair if its false positive rate and its false negative rate are each the same across all sub-populations. They proposed to quantify the deviation from 
equalized odds using a measure that we will henceforth call \unfairnessname\ (\UNF). \UNF\ is the fraction of the population for which the classifier predicts
with conditional prediction probabilities that differ from the closest common baseline.

The fairness notion of equalized odds was originally studied for binary classifiers. However, interest in fairness for \emph{multiclass} classifiers has gained traction in recent years \citep[see, e.g.][]{AlghamadiEtAl22, rouzot2022learning, wang2024aleatoric}. In multiclass scenarios, the multiclass equalized odds criterion measures any differences in conditional prediction probabilities between sub-populations. This includes not only the difference in the rate of correct predictions as in the binary case, but also the types of prediction mistakes. For instance, if a patient's heart attack is misdiagnosed as an anxiety attack (which may mean the patient is denied care), this is significantly different than being misdiagnosed as a stroke (which may lead to delayed care). If some sub-populations incur more of a certain type of misdiagnosis error, this could indicate bias in diagnosis, as well as lead to undesired differences in treatment.

In this work, we study the auditing of multiclass classifiers for deviation from multiclass equalized odds, using a natural generalization of \UNF.
\UNF\ is different from other commonly used measures of deviation from equalized odds, such as the difference or ratio of equalized odds \citep[e.g.,][]{AlghamadiEtAl22, wang2024aleatoric}, in that it has a consistent interpretable meaning as a fraction of the population, regardless of the number of protected attribute values, the number of classes, or the degree of class imbalance. Thus, \UNF\ is useful for interpretably auditing and comparing classifiers. This is contrasted with the standard use of differences or ratios, which produces undesirable artifacts, such as discounting differences in rare labels, lack of normalization or boundedness, and lack of differentiation between classifiers with different degrees of bias. The quantifiable interpretation of the \UNF\ measure ensures that it does not suffer from similar issues.

It was shown in \citet{SabatoYo20} that the \UNF\ of a given binary classifier can be calculated efficiently using the confusion matrices of this classifier in each sub-population. When the confusion matrices are not available as discussed above, there exists an efficient procedure for deriving a lower bound on the value of \UNF\ given population-level frequencies. This lower bound can be used to identify classifiers that treat a high fraction of the population unfairly, without direct access to the classifier. 
In the case of multiclass classifiers, efficient procedures for calculating \UNF\ are unknown, as this may be computationally intractable. Thus, we propose procedures to upper-bound and lower-bound the \UNF\ of a multiclass classifier, given the conditional confusion matrices of the classifier for each sub-population, as well as given only population-level frequencies. 

The upper bounds are obtained using local minimization procedures. The minimization problems are constrained, non-smooth and non-convex, and their objective functions have regions with large gradients, which is challenging numerically. As it is known that non-smooth functions are challenging for gradient-based algorithms, we first handle the non-smoothness, caused by a maximum term over several concave functions, by splitting the functions into their smooth parts, and adding more constraints to the existing ones to account for that. Then, we replace the concave functions in the constraints with linear approximations, and the minimization is obtained via sequential solutions of standard linear programming (LP) problems, which have available and rather efficient solution routines \citep{hall2023highs}. The LP solvers efficiently handle the constrained minimization at each step, even when the constraint matrices include large numbers because of the large gradients. As the problem is highly non-convex, the sequential minimization reaches local minima.

We report experiments on several data sets, showing that the gap between the
upper and lower bounds, for both scenarios, are usually quite small,
indicating that the optimization procedures provide useful estimates. These estimates can be used to identify classifiers that behave differently on different protected sub-populations.

\paragraph{Paper structure.}
\secref{related} discusses related work. Preliminaries are provided in \secref{settings}. In \secref{dcb}, we present the \UNF\ measure and extend it to multiclass classifiers. \secref{multiclassknown} provides methods for calculating upper and lower bounds for \UNF\ for multiclass classifiers when the conditional confusion matrices for all sub-populations are known.  In \secref{multunknown}, we consider the case where these matrices are unknown, using only population-level frequencies. Experiments are reported in \secref{exps}. We conclude in \secref{discussion}. Some technical details are deferred to appendices.

\section{Related Work}\label{sec:related}

Fairness for multiclass classification has been gaining interest in recent years. \citet{denis2024fairness} studied multiclass fairness with demographic parity.
\citet{AlghamadiEtAl22} use model projection for multiclass equalized odds. \citet{putzel2022blackbox} propose post-processing techniques for obtaining fairness in multiclass classification for various fairness notions. \citet{rouzot2022learning} propose fairness scoring systems for multiclass classification. \citet{wang2024aleatoric} study the fundamental limits of fairness in multiclass classifiers, under several fairness notions, including equalized odds.

\citet{SabatoYo20} studied estimating the (un)fairness of a classifier, a crucial task in many applications \citep{Bellamy19}. They proposed the new \UNF\ measure for binary classifiers, showed that it is easy to calculate using the confusion matrices for each sub-population, and provided methods for lower-bounding this measure in the absence of individual label data, using population-level frequencies. Several other works have studied fairness auditing in the absence of individual information about protected attributes \citep[e.g.,][]{awasthi2021evaluating, fabris2023measuring, cornacchia2023auditing}. Fairness auditing has also been studied in specific applications, including tax auditing algorithms \citep{BlackElChGoHo22}, visual systems \citep{GoyalSoHaSaUs22}, and candidate rankings \citep{RothSaYu22}. 

Many works use some relaxation of equalized odds to allow learning or studying near-fair classifiers. However, there is not a single agreed-upon relaxation, in the binary or the multiclass case. For instance, \citet{DoniniOnBeShPo18,jung2021fair,xue2023group,wang2024aleatoric} use a difference-based formulation, while \citet{Calmonetal2017} and \citet{AlghamadiEtAl22} use a ratio-based one. In \citet{xue2023group}, the sum of the differences is used, while in \citet{wang2024aleatoric}, the maximum is used. In this work, we study a natural extension to of the interpretable \UNF\ measure of \citet{SabatoYo20} to multiclass classification.

Our work partly falls within the realm of fairness auditing using only aggregate statistics, without assuming access to the classifier. The challenge of auditing fairness using limited information has received significant attention in recent years, as evident, for example, in \citep{pinzon2024incompatibility, wang2021fair}. Our work is unique, in that it is the first, to our knowledge, to address the multiclass setting.

\section{Preliminaries}\label{sec:settings}

We consider a multiclass classification problem with $k$ possible labels, in which each individual in the population has a true label in the label set $\cY \equiv \{1,\ldots,k\}$. Fairness is considered with respect to some protected attribute, such as race, state of residence or age. A value of the attribute, which can be multi-valued, is assigned to each individual. If there is more than one protected attribute, it can be substituted for a single attribute which is the Cartesian product of all protected attributes.
A \emph{sub-population} is the subset of the population that includes all the individuals with the same value of the protected attribute.

The object of study is an existing classifier, denote it $\cC$, which maps each individual from the population to a predicted label, which may be different from its true label. We do not make any assumptions about the way the classifier is generated or the classification model. For a given $\cC$, denote by $\cD$ the uniform distribution over the population of the triplets of true label, predicted label, and protected attribute value of individuals. A random triplet drawn according to $\cD$ is denoted by $(Y, \hat{Y}, A)$, where $Y \in \cY$ is the true label of the individual, $\hat{Y} \in \cY$ is the label predicted by $\cC$ for this individual, and $A \in \cA$ is the individual's protected attribute value, where $\cA$ is the set of possible values.
Denote the probability of an event $E$ according to $\cD$ by $\P[E]$.

We define the following notation for properties of $\cD$:
  The frequency of each sub-population in the distribution is $w_a := \P[A = a]$; The vector of frequencies is $\bw = (w_a)_{a \in \cA}$.
  The proportion of true label $y\in \cY$ in sub-population $a \in \cA$ is $\pos^y_a := \P[Y=y \mid A = a]$; The vector of these proportions is $\bpi_a := (\pi_a^y)_{y \in \cY}$. The proportion of predicted label $y \in \cY$ by classifier $\cC$ in sub-population $a \in \cA$ is $\hp^y_a := \P[\hat{Y} = y \mid A = a]$; The vector of these proportions is $\bhp_a := (\hp_a^y)_{y \in \cY}$.
  The entries in the confusion matrix of the classifier $\cC$ on a sub-population $a \in \cA$ are denoted by
$\forall y,\hat{y} \in \cY,  \alpha_a^{y\hat{y}} := \P[\hat{Y} = \hat{y} \mid Y = y, A = a].$
Throughout this work, the value of a conditional probability expression in which the probability of the condition is zero is treated as zero.
Denote the confusion matrix of $\cC$ on a sub-population $a$ by $\confg := (\alpha_a^{y\hat{y}})_{y,\hat{y}\in \cY}$. Denote the indexed set of all such confusion matrices by $\confgall := \{\confg\}_{a \in \cA}$. Denote row $y$ in $\confg$ by $\allag^y := (\alpha_a^{y\hat{y}})_{\hat{y}\in \cY}$.

Denote the $k$-simplex by $\Delta_k := \{ \bv \in \reals_+^{k} \mid \norm{\bv}_1 = 1\}$. The set of matrices whose rows are in the simplex over $\cY$ is $\matdelta := \{ \mathbf{X} := (x^{y\hat{y}})_{y,\hat{y} \in \cY} \mid \forall y \in \cY, (x^{y\hat{y}})_{\hat{y} \in \cY} \in \Delta_k\}$. 
By definition, the classifier $\cC$ satisfies the following:
\begin{align}\label{eq:fairconst}
        \forall a \in \cA \text{ it holds that } &\confg \in \matdelta \text{ and }\confg^T\bpi_a = \bhp_a.
\end{align}

The fairness notion of  \emph{equalized odds} \citep{HardtPrSr16}, originally defined for binary classification, was later generalized to multiclass classifiers. We use the term-by-term multiclass equalized odds criterion \citep{putzel2022blackbox, AlghamadiEtAl22}, which requires that for each $y,\hat{y} \in \cY$, $\alpha_a^{y\hat{y}} \equiv \P[\hat{Y}=\hat{y} \mid Y = y, A=a]$ is the same across all $a \in \cA$. Equivalently, all the matrices in $\bfM_\cA$ are the same.
Note that the same definitions can be used also for relaxed multiclass equalized odds criteria, such as those that distinguish sensitive and insensitive labels  \citep[see, e.g.][]{rouzot2022learning}. This can be achieved by mapping $\hat{Y}$ conditioned $Y = y$ to a smaller set of predicted labels, where the mapping can depend on $y$, and then recalculating the distribution properties provided above. All the methods and results below are valid for the resulting transformation.

\section{The \UNF\ Measure for Multiclass Classifiers}\label{sec:dcb}

\begin{figure}
  \begin{center}
\begin{subfigure}{0.7\columnwidth}
  \centering
  \includegraphics[width=.99\columnwidth]{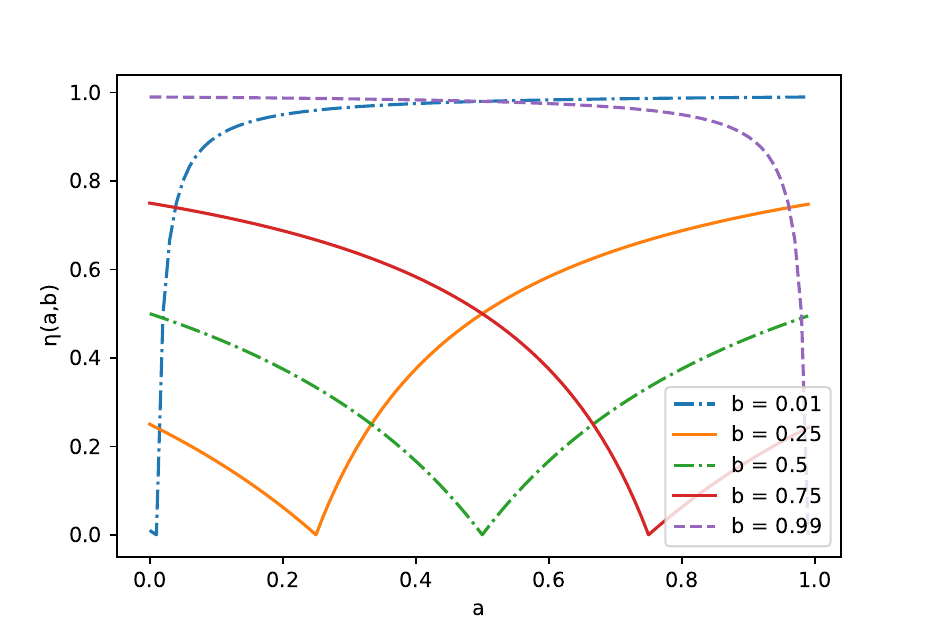}
  \caption{ $a \mapsto \eta(a,b)$}
\end{subfigure}
\begin{subfigure}{0.7\columnwidth}
  \centering
  \includegraphics[width=.99\columnwidth]{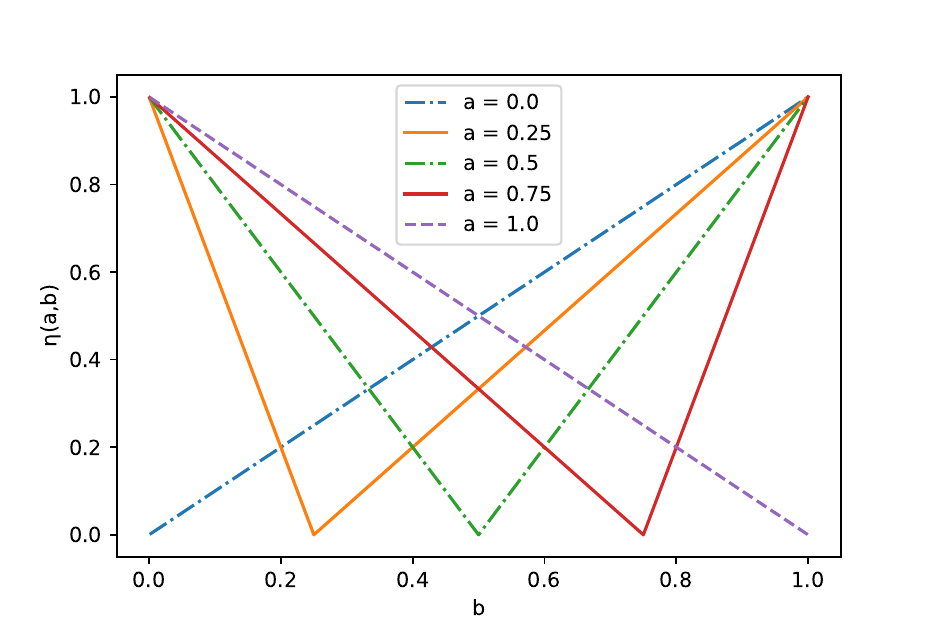}
  \caption{ $b \mapsto \eta(a,b)$}
  \label{fig:sfig2}
\end{subfigure}
\end{center}
\caption{The function $\eta(a,b)$.}
\label{fig:eta}
\end{figure}

Recall that the distribution $\cD$ is determined by the given classifier $\cC$. Denote the conditional distribution of $\cD$ given the attribute value $A=a$, and the true label $Y=y$ by $\cD^y_a$. The binary \UNF\ measure is based on modeling each $\cD_a^y$ as a mixture of two conditional distributions: a global baseline distribution, which is the same for all sub-populations $a$ and represents the baseline behavior, and a local nuisance distribution, which can be different for each $a$ and represents the deviation from this baseline. For $a \in \cA, y \in \cY$, let $\eta_a^y \in [0,1]$ be the probability conditioned on \mbox{$A=a, Y=y$} that $\hat{Y}$  is drawn according to the nuisance distribution. $\UNF(\cC)$ is defined as the fraction of the population for which the local nuisance conditional distributions are used by the classifier instead of the global baseline distribution. Since the baseline conditional distribution is unobserved, it is taken to be the one that results in the minimal possible \UNF. Formally, for a given classifier $\cC$ with the distribution $\cD$, and letting $\bfeta_a := (\eta_a^y)_{y \in \cY}$, 
\begin{equation}\label{eq:unfbasic}
  \mathrm{\UNF}(\cC) = \min_{\{\bfeta_a\}}\sum_{a \in \cA} w_a \bpi_a^T\bfeta_a,
\end{equation}
where the minimum is taken over ${\{\bfeta_a\}_{a \in \cA}}$ that are consistent with $\cD$; that is, such that there exist a baseline distribution $\dbase^y$ and nuisance distributions $\{\cN_a^y\}_{y \in \cY, a \in \cA}$ such that for each $y \in \cY, a \in \cA$, $\cD_a^y = (1-\eta^y_a)\dbase^y  + \eta_a^y \cN^y_a$. If the classifier satisfies equalized odds, then the distributions $\cD_a^y$ are the same for all $a \in \cA$, in which case the decomposition holds by setting $\eta_a^y = 0$ for all $y \in \cY, a \in \cA$ and setting the baseline distribution for $y$ to $\cD_a^y$ for an arbitrary $a$. This gives $\UNF(\cC) = 0$, as expected.
\citet{SabatoYo20} show that for binary classifiers, where $\cY = \{0,1\}$,   
\begin{align}\label{eq:unfbinary}
  \mathrm{\UNF}&(\cC) = \mathrm{\UNF}(\confgall, \bw, \bpi) := \\
   & \sum_{y \in \cY} \min_{\alphabase^{y(1-y)} \in [0,1]}\sum_{a \in \cA} w_a  \pos_a^y \eta(\alphabase^{y(1-y)},\alpha_a^{y(1-y)}),\nonumber
\end{align}
where $\confgall$ is determined by $\cC$, $\bw, \bpi$ are the population properties, and $\eta$ without subscripts denotes the function:
\begin{equation}\label{eq:etadef}
\eta(a, b) = \begin{cases}
1-b/a & b < a,\\ 
1-(1-b)/(1-a) & b > a,\\
0 & b=a.
\end{cases}
\end{equation}
$\eta$ is illustrated in Figure \ref{fig:eta}. In the minimization in \eqref{unfbinary}, $\alphabase^{y(1-y)}$ represent conditional prediction rates for the baseline distribution. More generally, we denote the conditional probability of predicting $\hat{y}$ given true label $y$ under the baseline distribution by $\alphabase^{y\hat{y}}$. \citet{SabatoYo20} observe that since $a \mapsto \eta(a,b)$ is piecewise concave on the intervals $[0,b]$ and $[b,1]$ (see \figref{eta}), it suffices to minimize each of the terms in \eqref{unfbinary} over $\alphabase^{y(1-y)} \in \{\alpha_a^y\}_{a \in \cA} \cup \{0,1\}$.

We first provide a formulation of $\UNF(\confgall, \bw, \bpi)$ for multiclass classifiers.

\begin{theorem} For multiclass classification, \eqref{unfbasic} implies 
\begin{align}\label{eq:unfairmin}
    \UNF(\cC) = &\; \UNF(\confgall, \bw, \bpi) =\\
    & \sum_{y\in\cY}\min_{\alla^y \in \Delta_k}\sum_{a \in \cA} w_a \pi_a^y\max_{\hat{y} \in \cY}\eta(\alphabase^{y\hat{y}},\ta^{y\hat{y}}),\nonumber
\end{align}
where $\alla^y := (\alphabase^{y\hat{y}})_{\hat{y}\in \cY}$.
\end{theorem}
\begin{proof}
In an analog to the derivation in \citet{SabatoYo20},
it is easy to see that for a given baseline distribution with rates $\{\alphabase^{y\hat{y}}\}_{y,\hat{y}\in\cY}$,
\[
  \ta^{y\hat{y}} = \alphabase^{y\hat{y}}(1-\eta^y_a) + \nu_a^{y\hat{y}}\cdot \eta^y_a,
  \]
  where $\nu_a^{y\hat{y}}$ is the conditional probability of predicting $\hat{y}$ given true label $y$, under the nuisance distribution $\cN_a^y$.
  From \eqref{unfbasic}, it follows that the optimal solution for $\nu_a^{y\hat{y}}$ minimizes $\eta^y_a$ subject to
  \begin{align*}
    &\forall \hat{y} \in \cY \st \nu_a^{y\hat{y}} \neq \alphabase^{y\hat{y}}, \eta_a^y = \frac{\ta^{y\hat{y}}-\alphabase^{y\hat{y}}}{\nu_a^{y\hat{y}}-\alphabase^{y\hat{y}}}, \\
    &\forall y \in \cY, a \in \cA, \eta_a^y \geq 0,\notag \\
    \notag
    &\forall a \in \cA, (\nu_a^{y1},\ldots,\nu_a^{yk}) \in \Delta_k.
  \end{align*}

  If $\alpha_a^{y\hat{y}} = \alphabase^{y\hat{y}}$ for all $\hat{y} \in \cY$, then clearly $\eta_a^y = 0$. Otherwise, the first constraint requires $\nu_a^{y\hat{y}} =  \alphabase^{y\hat{y}}+(\ta^{y\hat{y}}-\alphabase^{y\hat{y}})/\eta_a^y$. Summing over $\hat{y}$, this implies
  $\sum_{\hat{y} \in \cY} \nu_a^{y\hat{y}} =  \sum_{\hat{y} \in \cY}\alphabase^{y\hat{y}}+(\sum_{\hat{y} \in \cY}\ta^{y\hat{y}}-\sum_{\hat{y} \in \cY}\alphabase^{y\hat{y}})/\eta_a^y = 1$. Thus, the last constraint can be replaced by $\nu_a^{y\hat{y}} \in [0,1]$ for all $\hat{y} \in \cY$. 
  Since the derivative of the RHS of the first constraint is never zero, the minimizer of $\eta_a^y$ is obtained when one of the other constraints holds. That is, with an assignment of $\nu_a^{\hat{y}y} = 1$ for some $a,\hat{y},y$ such that $\alpha_a^{y\hat{y}} > \alphabase^{y\hat{y}}$, or $\nu_a^{\hat{y}y} = 0$ for some $a,\hat{y},y$ such that $\alpha_a^{y\hat{y}} < \alphabase^{y\hat{y}}$.
  It follows that there is some $\hat{y} \in \cY$ such that
    \begin{align*}
    &\ta^{y\hat{y}} > \alphabase^{y\hat{y}} \text{ and } \eta^y_a = \frac{\ta^{y\hat{y}}-\alphabase^{y\hat{y}}}{1-\alphabase^{y\hat{y}}} = 1-\frac{1-\ta^{y\hat{y}}}{1-\alphabase^{y\hat{y}}}\\
    &\qquad\qquad\qquad\text{ or }\\
    &\ta^{y\hat{y}} < \alphabase^{y\hat{y}} \text{ and }\eta^y_a = \frac{\ta^{y\hat{y}}-\alphabase^{y\hat{y}}}{-\alphabase^{y\hat{y}}} = 1-\frac{\ta^{y\hat{y}}}{\alphabase^{y\hat{y}}}.
  \end{align*}
  Moreover, for all $\hat{y} \in \cY$, if $\ta^{y\hat{y}} > \alphabase^{y\hat{y}}$ then $\eta_a \geq 1-\frac{1-\ta^{y\hat{y}}}{1-\alphabase^{y\hat{y}}}$ and if $\ta^{y\hat{y}} < \alphabase^{y\hat{y}}$ then $\eta_a \geq 1-\frac{\ta^{y\hat{y}}}{\alphabase^{y\hat{y}}}$.
  It follows that 
  \begin{equation*}
  \eta^y_a = \max_{\hat{y} \in \cY} \max(1-\frac{1-\ta^{y\hat{y}}}{1-\alphabase^{y\hat{y}}}, 1-\frac{\ta^{y\hat{y}}}{\alphabase^{y\hat{y}}}) = \max_{\hat{y} \in \cY}\eta(\alphabase^{y\hat{y}},\ta^{y\hat{y}}).
  \end{equation*} 
  \eqref{unfairmin} thus follows.
 \end{proof}
 It is easy to verify that \eqref{unfairmin} is equivalent to \eqref{unfbinary} for binary classifiers, since for $\cY = \{0,1\}$, $\alpha_a^{yy} = 1-\alpha_a^{y(1-y)}$ and the same for $\alphabase$, and $\eta(a,b) = \eta(1-a,1-b)$.

  For simplicity of notation, we henceforth treat $\bw$ and $\bpi$ as fixed and write $\UNF(\confgall)$. 
  In the multiclass case, $\UNF(\confgall)$ is the solution of a non-convex constrained minimization problem. In the next section, we provide methods for calculating a lower bound and an upper bound for this quantity.

Note that in practice, in some cases, the properties of the distribution $\cD$ for $\cC$ would be estimated from a limited data set and so may deviate from the true properties for $\cD$. However, since $\UNF$ is interpretable as a fraction of the population, any inaccuracy in the estimation of $\cD$ would map to at most the same amount of inaccuracy in $\UNF$.

\section{Bounding the \UNF\ of a Multiclass Classifier}\label{sec:multiclassknown}
We propose methods for calculating a lower bound and an upper bound for $\UNF(\confgall)$. The lower bound is a simple analytical formula.
  For $y \in \cY$, define $\allabigg^y := \{ \alpha_a^y \}_{a \in \cA}$ and 
  \begin{equation}\label{eq:miny}
    \UNF_y(\allabigg^y) := \min_{\alla^y \in \Delta_k}\sum_{a \in \cA} w_a \pi_a^y\max_{z \in \cY}\eta(\alphabase^{y\hat{y}},\ta^{y\hat{y}}).
  \end{equation}
  Then $\UNF(\confgall) = \sum_{y \in \cY} \UNF_y(\allabigg^y)$. 
  A lower bound on $\UNF_y$ can be derived as follows:
\begin{align}\label{eq:unfairlower}
  \UNF_y(\allabigg^y) &= \min_{\alla^y \in \Delta_k}\sum_{a \in \cA} w_a \pi_a^y\max_{\hat{y} \in \cY}\eta(\alphabase^{y\hat{y}},\ta^{y\hat{y}})\\
  &\geq  \min_{\alla^y \in [0,1]^{k}}\sum_{a \in \cA} w_a \pi_a^y\max_{\hat{y} \in \cY}\eta(\alphabase^{y\hat{y}},\ta^{y\hat{y}})&\notag\\
  &\geq\min_{\alla^y \in [0,1]^{k}}\max_{\hat{y} \in \cY}\sum_{a \in \cA} w_a \pi_a^y\eta(\alphabase^{y\hat{y}},\ta^{y\hat{y}})&\notag\\
                      &= \max_{\hat{y} \in \cY}\min_{x \in [0,1]}\sum_{a \in \cA} w_a \pi_a^y\eta(x,\ta^{y\hat{y}}).\notag
\end{align}
Due to the piecewise concavity of $a \mapsto \eta(a,b)$ on $[0,b]$ and $[b,1]$, it suffices to minimize each of the terms in \eqref{unfbinary} over $x \in \{0,1\}\cup\{\alpha_a^{y\hat{y}}\}_{a \in \cA}$. Thus,
\[
  \UNF(\confgall) \geq \sum_{y\in \cY}\max_{\hat{y} \in \cY}\min_{x \in \{0,1\}\cup\{\alpha_a^{y\hat{y}}\}_{a \in \cA}}\sum_{a \in \cA} w_a \pi_a^y\eta(x,\ta^{y\hat{y}}).
\]
This lower bound can be calculated in time $O(|\cA|^2k^2)$.

To upper bound $\UNF(\confgall)$, we propose a local iterative optimization approach on the objective function in \eqref{miny}, which is constrained, non-smooth, and non-convex. One can randomly initialize using a feasible solution. However, we propose a more tailored approach at the end of this section.

Our optimization approach utilizes sequential LP solutions, and is as follows. Given $y \in \cY$, denote
the matrix $\bfH\in\mathbb{R}^{|\cA|\times k}$ to include the entries $h_{a,\hat{y}} := \alpha_a^{y\hat{y}}$. Note that the rows of $\bfH$ are in the simplex $\Delta_k$. 
  The vector $\bfw\in[0,1]^{|\cA|}$ includes the entries $\tw_a := w_a \pi_a^y$.
  The optimization variables are denoted by the vector $\bfalpha := \alla^y \in\mathbb{R}^{k}$.
  Denote the all-one vector of dimension $d$ by ${\bf 1}_d$ and the identity matrix of dimensions $d\times d$ by $\mathbf{I}_d$.
Our objective in \eqref{unfairmin} and its constraints, is thus given by:
\begin{alignat}{2}\label{eq:obj1}
&\Minimize_{\bfalpha\in\mathbb{R}^{k}}~~  &&\sum_a \tw_a \max_{\hat{y}} \{\eta(\alpha_{\hat{y}}, h_{a,\hat{y}})\} \\
&\st && 0 \leq \bfalpha \leq 1, \nonumber \\
&&&  \langle \bfalpha, {\bf 1}_k\rangle=1. \nonumber
\end{alignat}
To solve the problem, we observe that its structure resembles a linear program (LP). That is, the objective is given by an inner product, and the constraints are linear. The only difference between an LP and \eqref{obj1} is the maximum term over the $\eta$ values inside the inner product. Hence, to solve \eqref{obj1}, we apply the sequential linear programming approach \citep{nocedal2006numerical}, which iteratively approximates \eqref{obj1} as a linear program and solves it.

First, we replace the maximum terms in the objective with another variable vector $\bfc$ and additional constraints, following \citet{charalambous1978efficient}:
\begin{alignat}{2}\label{eq:obj_constraints}
&\Minimize_{\bfalpha\in\mathbb{R}^{k}, \bfc\in\mathbb{R}^{|\cA|}}~~ &&\langle \bfw, \bfc\rangle \\
&\st && 0 \leq \bfalpha, \bfc \leq 1, \nonumber \\
&&&  \langle \bfalpha, {\bf 1}_k\rangle=1, \nonumber\\
&&&  \eta(\alpha_{\hat{y}}, h_{a,\hat{y}})\leq c_a, \forall \hat{y} \in [k], \forall a \in \cA.\nonumber
\end{alignat}
This problem is equivalent to \eqref{obj1}, but has no maximum operation. However, now we have non-linear constraints, thus violating the definition of an LP. To correct this, we use a local approximation of $\eta$ (around an iterate $\bfalpha^{(t)}$) using the Taylor series. Note that the function $a \mapsto \eta(a,b)$ is piecewise concave, non-negative, and smooth at $(0,1)$ except at $b$, where it is minimized and equals zero. Therefore, we do not expect the maximum over $\hat{y}$ in the objective \eqref{obj1} to fall on the discontinuity of $\eta$ at $b$, except for extreme cases. Hence, using a linear approximation we expect to get a good local approximation of $\eta()$.

\begin{algorithm}[t]
  \begin{algorithmic}
\STATE\textbf{Input: } $\bfalpha^{(0)}, \bfH, \bfw,\mathrm{maxIter}, \varepsilon$
\STATE\textbf{Output:} The local minimizer $\bfalpha^{*}$
\FOR{$t = 0$ \textbf{to} $\mathrm{maxIter}$}
\STATE Define the LP approximation of \eqref{obj_constraints} by computing $\bfM_1, \bfM_2$, and $\bfv$.
\STATE Define $\hat \bfalpha$ as the minimizer of the LP in \eqref{obj_constraints_taylor_canonical}.
\STATE Set $\bfalpha^{(t+1)} = \bfalpha^{(t)} + \mu(\hat \bfalpha - \bfalpha^{(t)}),$
  where $\mu$ is obtained using line search to ensure a reduction in \eqref{obj1}.
  \STATE If $\|\bfe\| < \varepsilon$, stop.
\ENDFOR
\STATE \textbf{return} the last iterate $\bfalpha^{(t)}$.
\end{algorithmic}
\caption{Local minimization of $\UNF_y$\ via sequential linear programming}
\label{alg:FairV1}
\end{algorithm}

The sequential linear programming approach calculates the new iterate
$(\bfalpha^{(t+1)},\bfc^{(t+1)})$ by solving an LP problem of an approximation of \eqref{obj_constraints} around $(\bfalpha^{(t)},\bfc^{(t)})$.
That is, given an iterate $(\bfalpha^{(t)},\bfc^{(t)})$ we first approximate $\eta$ by a linear Taylor series in its first argument:
\begin{equation*}
\eta(\alpha+\epsilon,b) - \eta(\alpha,b) \approx  \frac{\partial\eta(\alpha,b)}{\partial \alpha}\epsilon =  \left\{\begin{array}{lc}
\frac{b}{\alpha^2}\epsilon  & \alpha>b,\\
\frac{-(1-b)}{(1-\alpha)^2}\epsilon & \alpha<b,\\
0 & \alpha=b.\end{array}\right.
\end{equation*}
We now locally approximate \eqref{obj_constraints} around an iterate $\bfalpha^{(t)}$ via the following LP problem:
\begin{alignat}{2}\label{eq:obj_constraints_taylor}
&\Minimize_{\bfalpha\in\mathbb{R}^{k},\bfc\in\mathbb{R}^{|\cA|}}\quad&& \langle \bfw, \bfc\rangle \\
& \quad \st &&  0 \leq \bfalpha,\bfc  \leq 1, \nonumber \\
&&&   \langle \bfalpha, {\bf 1}_k\rangle=1, \nonumber\\
&&&  \bfv_a + \bfJ_a(\bfalpha-\bfalpha^{(t)}) \leq {\bf 1}_{k}c_a, \forall a \in \cA,\nonumber
\end{alignat}
where $\bfv_a \in \reals^{k}$ is a vector with the coordinates $(\bfv_a)_{\hat{y}} = \eta(a_{\hat{y}}^{(t)}, h_{a,\hat{y}})$ for all $\hat{y} \in [k]$, and $\bfJ_a\in\mathbb{R}^{k\times k}$ is a diagonal Jacobian matrix s.t $(\bfJ_a)_{\hat{y}\hat{y}} = \frac{\partial\eta(a_{\hat{y}}^{(t)}, h_{a,\hat{y}})}{\partial a}$.
Bringing this into a canonical form with a variable $\bfalpha = \bfalpha^{(t)}+\bfe$ yields
\begin{alignat}{2}\label{eq:obj_constraints_taylor_canonical}
&\Minimize_{\bfalpha\in\mathbb{R}^{k},\bfc\in\mathbb{R}^{|\cA|}} \quad&&\langle \bfw, \bfc\rangle \\
&\quad \st&& 0 \leq \bfalpha,\bfc  \leq 1, \nonumber \\
&&&  \langle \bfalpha, {\bf 1}_k\rangle=1, \nonumber\\
&&&  \bfM_1\bfc + \bfM_2\bfalpha - \bfM_2\bfalpha^{(t)} + \bfv \leq 0,\nonumber
\end{alignat}
where $\bfM_1 := -\mathbf{I}_{|\cA|}\otimes {\bf 1}_{k}$, $\bfM_2 := [\bfJ_1;\bfJ_2;...;\bfJ_{|\cA|}]$ is the stacking of all diagonal Jacobian matrices on top of each other, and $\bfv := [\bfv_1;...;\bfv_{|\cA|}]$ is the stacking of vectors. \eqref{obj_constraints_taylor_canonical} is an LP in a canonical form, which is solved iteratively. Note that $\bfM_1$ and $\bfM_2$ are highly sparse, which can be exploited to easily solve large instances of the problem. The resulting local optimization algorithm is given in \myalgref{FairV1}.

\textbf{A note on usage.} By definition, the function $\eta$ in \eqref{etadef} is continuous in the open section $(0,1)^2$, but at the boundaries, it is easy to see that $\eta(0,0)=0$, while $\lim_{\delta\rightarrow 0}\eta(\delta,0) = 1$. Furthermore, the derivatives of $\eta$ when its first argument is close to $0$ or $1$ can be very large, causing numerical difficulties in the LP solvers. Thus, to avoid numerical instability, we make sure that all entries of $\bfH$ are in $[\varepsilon, 1-\varepsilon]$ (where we used $\varepsilon = 10^{-5}$) by replacing any value outside these boundaries by the relevant boundary value and renormalizing. Note that since the input values in $\bfH$ are calculated in practice based on  finite data sets, values in $\bfH$ are already noisy representations of the true population values, and so slightly changing them does not negatively affect the correctness of the method.

\textbf{Finding an initializing assignment} A natural guess for an initializing assignment $\bfalpha^{(0)}$ is the weighted average of the confusion matrices in $\confgall$. However, this does not take the objective function into account. We propose instead a greedy approach, in which the entries in the baseline confusion matrix are optimized label-by-label, using the fact that the binary problem is easy to minimize. In the first iteration, all labels except for one are treated as the same label, and optimal confusion matrix values for the first labels are calculated. In each further iteration, one additional label is separated, and the assignment for this label is optimized given the assignments of the previous labels. This approach is possible because \UNF\ is monotonic under the merging of labels. The full derivation and procedure are provided in \appref{greedy}. Our experiments in \secref{exps} demonstrate that this method is superior to the averaging approach.

\section{Best-case \UNF\ without Confusion Matrices}\label{sec:multunknown}

We next consider the case of a classifier for which the confusion matrices for each sub-population are not available, either because the classifier is not accessible for testing or because individual-level ground-truth data is missing or insufficient. In this case, we assume, as in \citet{SabatoYo20}, access only to the frequencies of true and predicted labels in each sub-population, but not the conditional probabilities. Formally, we do not have $\confgall$ for the given classifier $\cC$. We only have access to the population-level frequencies $(\bw, \{(\bpi_a,\bhp_a)\}_{a \in \cA})$.

If $\cC$ is fair under multiclass equalized odds, then all the matrices in $\confgall$ must be identical. Thus, from \eqref{fairconst}, there  exists a single confusion matrix $\bfM \in \matdelta$ such that $\forall a \in \cA, \bfM^T \bpi_a = \bhp_a.$ If such a matrix does not exist and the input frequencies are accurate, then $\cC$ does not satisfy the multiclass equalized odds criterion. Nonetheless, given $(\bw, \{(\bpi_a,\bhp_a)\}_{a \in \cA})$ for $\cC$, we would like to find the best-case value of \UNF\ for  $\cC$, denoted $\minunf$. Formally,
\begin{align}
\label{eq:minunf}
    &\minunf((\bw, \{(\bpi_a,\bhp_a)\}_{a \in \cA})) := \\
    & \min\{ \UNF(\confgall) \mid \forall a \in \cA, \confg\in \matdelta, \confg^T\bpi_a = \bhp_a\}.\nonumber
\end{align}

\citet{SabatoYo20} show that $\minunf$ can be calculated exactly for binary classifiers. In the multiclass case, we are not aware of an efficient algorithm for calculating the exact value of $\minunf$. Instead, we provide below a local-minimization algorithm for $\UNF(\confgall)$ under the constraints on $\confgall$, which results in an upper bound for $\minunf$. This, combined with a lower bound on $\minunf$, provides a limited range of possible values of $\minunf$ for the given classifier $\cC$.

To calculate the lower bound, observe that
\begin{align}\label{eq:miny2}
    & \UNF(\confgall) = \sum_{y \in \cY} \min_{\alla^y \in \Delta_k}\sum_{a \in \cA} w_a \pi_a^y\max_{z \in \cY}\eta(\alphabase^{y\hat{y}},\ta^{y\hat{y}}) \nonumber \\
    & \geq \sum_{y \in \cY}\min_{\alphabase^{yy} \in [0,1]}\sum_{a \in \cA} w_a \pi_a^y\eta(\alphabase^{yy},\ta^{yy}).
  \end{align}
  Each of the terms in the RHS is of the same form as the terms in the definition of $\UNF$ for binary classification. The RHS is a constrained minimization problem that can be solved using the same methodology as in \citet{SabatoYo20}, providing a lower bound for $\minunf$.

  We now turn to the local optimization algorithm for $\minunf$.
We use a sequential linear programming approach, similarly to our solution to \eqref{obj1} in \secref{multiclassknown}. First, we extend the notation of \eqref{obj1} to explicitly denote the value of $y\in\mathcal{Y}$:
  The matrix  $\bfH^y \in\mathbb{R}^{|\cA|\times k}$ is defined such that $h^y_{a\hat{y}} = \alpha^{y\hat{y}}_a$;
  The vector $\bfw^y\in[0,1]^{|\cA|}$ includes the entries $\tw_a^y := w_a \pi_a^y$;
  The optimization variables are denoted by the vectors $\bfalpha^y := \alla^y \in\mathbb{R}^{k}$.
  We replace the maximum terms in the problem in \eqref{unfairmin} with the vectors $\bfc^y \in \mathbb{R}^{k}$, similarly to the transformation from \eqref{obj1} to \eqref{obj_constraints}. 
The variables for the optimization problem are the collection of the vectors and matrices
\begin{equation}\label{eq:x_multiclass}
\bfx = \{\bfalpha^1,\bfH^1,\bfc^1,...,\bfalpha^k,\bfH^k,\bfc^k\},
\end{equation}
which amount to $k^2+k^2\cdot|\mathcal{A}| + k\cdot|\mathcal{A}|$ scalar unknowns in total. 
We get the following new problem, which is equivalent to the problem defined in \eqref{unfairmin}, 
including its constraints: 
\begin{alignat}{2}\label{eq:obj_constraints_Large}
&\Minimize_{
\{\bfalpha^y\},\{\bfH^y\},\{\bfc^y\}
}
\quad &&\sum_y \langle \bfw^y, \bfc^y\rangle \\
&\quad \st  && 0 \leq \bfalpha^y \leq 1 \quad\forall y\in[k], \nonumber \\
&&& 0 \leq \bfH^y \leq 1 \quad\forall y\in[k], \nonumber \\
&&& 0 \leq \bfc^y \leq 1 \quad\forall y\in[k], \nonumber \\
&&&   \langle \bfalpha^y, 1\rangle=1 \quad\forall y\in[k], \nonumber\\
&&&   \bfH^y \mathbf{1}_k = \mathbf{1}_{|\mathcal{A}|} \quad\forall y\in[k],\nonumber\\
&&&  \textstyle \sum_{y=1}^k \pi^y_ah^y_{a,\hat{y}} = p_a^{\hat{y}} \quad\forall \hat{y}\in[k], a\in[|\mathcal{A}|]. \nonumber\\
&&&  \eta(\alpha_{\hat{y}}^y, h^y_{a,\hat{y}})\leq c_a^y\quad \forall y,\hat{y}\in[k], \;a\in[|\mathcal{A}|],\nonumber
\end{alignat}
Note that the objective in \eqref{obj_constraints_Large} is linear, and its constraints are linear as well, except the bottom ones involving $\eta()$, similarly to \eqref{obj_constraints}. Also note that all of the linear constraints are separable in $y$, except for the last one, which introduces a coupling between the $\bfH^y$ of different $y$'s.

\begin{figure}
  \centering
  \includegraphics[width=\columnwidth]{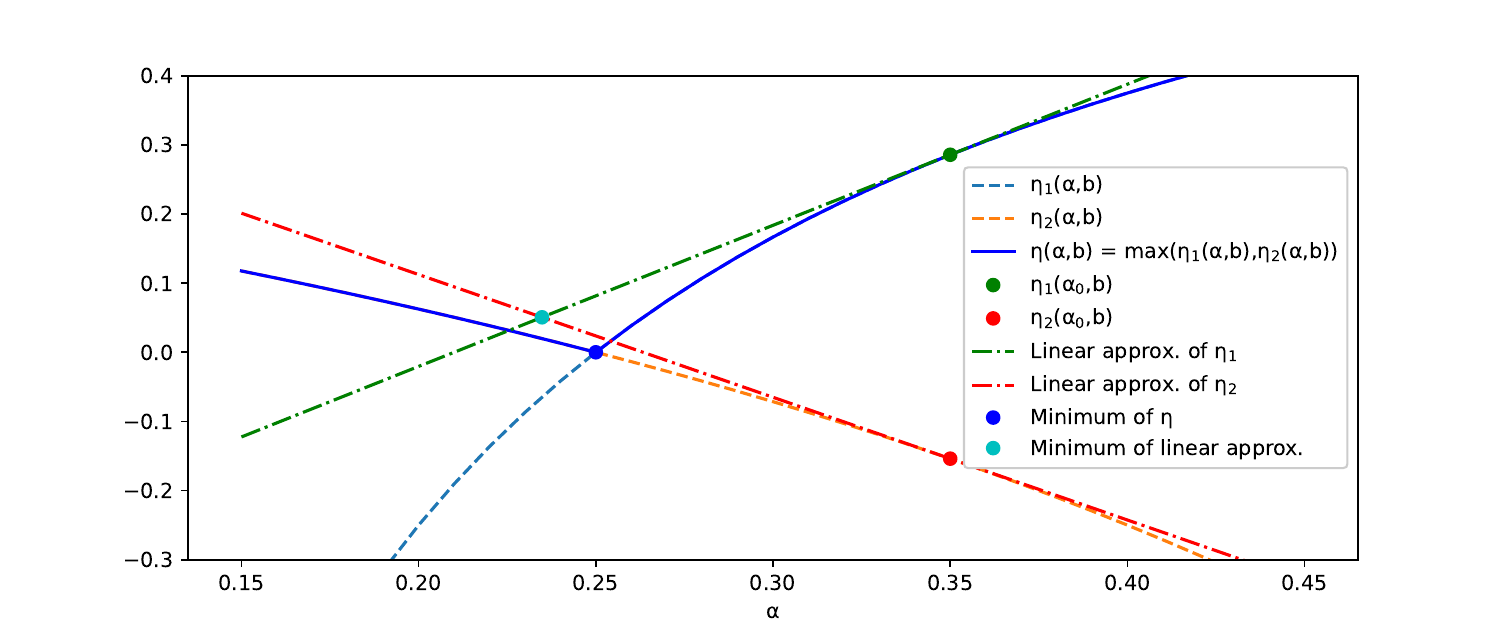}
  \caption{The linear approximation of $\eta(\alpha,b)$ using the split to $\eta_1$ and $\eta_2$ around $\alpha_0=0.35$, for a fixed $b=0.25$. }
  \label{fig:TaylorEta}
\end{figure}

To solve the problem we again use the sequential linear programming approach, as the structure of the problem again resembles a linear program. Given the $t$-th iterate $\bfx^{(t)}$ for all the variables defined in \eqref{x_multiclass}, we first approximate $\eta$ by a linear Taylor series, this time in both of its arguments, since now both are optimized. The function $\eta(\alpha,b)$ is linear in $b$, hence the Taylor series is exact with respect to $b$, which is appealing for our approach. However, minimizing the objective with respect to both $h^y_{a,\hat{y}}$ and $\alpha_a^y$ leads to the minimal non-smooth points of $\eta$. These points are the singularity points (see Fig. \ref{fig:eta}), where $\eta$ is not approximated well by a linear function. This results in a slow convergence of the sequential LP algorithm. To solve this, we split $\eta$ and define it as the maximum between the two smooth functions:
\begin{equation}\label{eq:splitEta}
    \eta(\alpha,b) = \max\left\{\eta_1(\alpha,b),\eta_2(\alpha,b)\right\},
\end{equation}
where $\eta_1(\alpha,b) = 1-\frac{b}{\alpha}$ and $\eta_2(\alpha,b) = 1-\frac{1-b}{1-\alpha}$.
This split is an alternative formulation to the definition of $\eta$ in \eqref{etadef}, only now the singularity point is replaced with a maximum, and can be treated in the same way we treat the maximization over all terms involving $\eta$ using sequential LP. Figure \ref{fig:TaylorEta} shows the independent linearization of the two terms after the split and the piece-wise linear approximation of the non-smooth function in our objective. Overall, the solution of the \eqref{obj_constraints_Large} is obtained after the split using sequential LP. More details are given in Appendix \ref{app:OPT2}.

\section{Experiments}\label{sec:exps}

We report experiments demonstrating the performance of the methods proposed above, for the case of known as well as unknown confusion matrices (\secref{multknownexp} and \secref{multunknownexp}), respectively. We show that in most cases, the ratio between the lower bound and the upper bound, in both types of experiments, is close to $1$, indicating that the methods produce fairly accurate estimates. Code is provided at \githublink.

\subsection{Bounding the \UNF\  with Known Confusion Matrices}\label{sec:multknownexp}

We report the results of calculating an upper bound and a lower bound on the $\UNF(\confgall)$ of a multiclass classifier given $\confgall$. As discussed above, if the confusion matrices in $\confgall$ are known up to some accuracy of $\cD$, then $\UNF(\confgall)$ will have at least the same accuracy, as a direct result of its definition as a fraction of the population.

We compared our proposed local-optimization approach for calculating an upper bound, described in \secref{multiclassknown}, to other approaches for finding a feasible low-value assignment for \eqref{unfairmin}. We tested the following alternatives: (1) Average: the weighted average of the sub-population confusion matrices; (2) Greedy: the greedy initialization procedure proposed in \secref{multiclassknown}; (3) Average+LM and (4) Greedy+LM: initializing with the corresponding method, then running the local optimization procedure.  
Our results below show that our approach, Greedy+LM, has the best performance.
We also report the ratio between the best upper bound and the lower bound, calculated according to \eqref{unfairlower}. A ratio close to $1$ indicates that in this case, the two bounds are tight.

For these experiments, we selected data sets on individuals, which are mostly categorical, and which include several multi-valued attributes that can be used as a predicted label and as a protected attribute. First, we used the US Census data set \citep{UCI2019} to generate multiclass classifiers to predict each of the individual attributes that have between $3$ and $10$ values. The protected attribute was the state of work. We tested our methods with three different types of classifiers: a decision tree and a nearest neighbor classifier, using standard Matlab libraries, as well as a classifier based on a standard fully connected neural network. The latter used two layers of size 64 and 32 neurons, with ReLU activation. The network was implemented in PyTorch, and trained with the AdamW optimizer, with a learning rate of 0.001, and a batch size of 256, for 4 epochs.

\tabref{censusdec},  as well as \tabref{censusnn} and \tabref{censusmlp} in \appref{tables}, list the $\UNF$ lower bound calculated for each of these classifiers given their confusion matrices, the upper bound obtained by each of the tested upper bound methods, and the ratio between the smallest upper bound and the lower bound. In some of the experiments, the lower bound is equal to the upper bound, giving the exact value of the \UNF\ of this classifier. In all of the experiments except for a single case, our proposed approach provides the tightest upper bound. The ratio between the upper bound and the lower bound ranges from $1.00$ to $2.85$, and is close to $1.00$ in most cases, showing that for these classifiers, the \UNF\ can be estimated quite well. In most cases, the $\UNF$ for decision tree classifiers obtained better accuracy ratios.

\begin{table}[h]
  \caption{\small \UNF\ with known confusion matrices, decision trees; US Census (top), Natality (bottom). For US Census, each row corresponds to a different classifier. Its error is indicated in the ``Error'' column. ``Lower bound'' is the lower bound calculated for DCP using our method. The upper bounds are the values obtained by each of the compared methods for minimizing the DCP objective. ``Best ratio'' indicates the ratio between the best (lowest) upper bound and the lower bound. A ratio of $1$ indicates that the bounds are both tight. For Natality, all the rows report the same classifier, and each row calculates DCP with respect to a different protected attribute.
  }
  \label{tab:censusdec}
   \resizebox{.5\textwidth}{!}{% <------ Don't forget this %
  \centering
  \begin{tabular}{c|c|c|cccc|c}
    \hline
    \# Labels & Error & Lower  & \multicolumn{4}{c|}{Upper Bounds} & Best \\
              & & Bound & Average & Greedy & Average$+$LM &  Greedy$+$LM &  Ratio \\
    \hline
3 &     $11.74\%$ & $5.39\%$ &$27.19\%$ & $9.65\%$ & $14.07\%$ & $\textbf{7.65\%}$ & $1.42$\\
3 & $5.71\%$ & $4.35\%$ &$42.17\%$ & $5.92\%$ & $32.39\%$ & $\textbf{5.28\%}$ & $1.21$\\
3 & $3.96\%$ & $3.24\%$ &$43.63\%$ & $5.07\%$ & $16.95\%$ & $\textbf{3.25\%}$ & $1.00$\\
3 & $5.15\%$ & $4.24\%$ &$39.05\%$ & $5.40\%$ & $14.32\%$ & $\textbf{4.25\%}$ & $1.00$\\
3 & $3.36\%$ & $2.65\%$ &$48.04\%$ & $5.20\%$ & $5.49\%$ & $\textbf{3.81\%}$ & $1.44$\\
3 & $1.85\%$ & $1.85\%$ &$59.64\%$ & $8.22\%$ & $17.85\%$ & $\textbf{1.85\%}$ & $1.00$\\
3 & $1.96\%$ & $1.96\%$ &$51.86\%$ & $7.88\%$ & $13.31\%$ & $\textbf{1.96\%}$ & $1.00$\\
3 & $2.32\%$ & $2.28\%$ &$48.50\%$ & $6.57\%$ & $10.56\%$ & $\textbf{2.28\%}$ & $1.00$\\
3 & $14.00\%$ & $4.57\%$ &$28.85\%$ & $6.47\%$ & $13.13\%$ & $\textbf{6.10\%}$ & $1.34$\\
 4 & $3.91\%$ & $1.48\%$ &$27.55\%$ & $8.12\%$ & $1.86\%$ & $\textbf{1.49\%}$ & $1.00$\\
    5 & $5.61\%$ & $2.06\%$ &$7.14\%$ & $43.01\%$ & $6.61\%$ & $\textbf{3.91\%}$ & $1.90$\\
5 & $11.83\%$ & $4.57\%$ &$25.28\%$ & $34.40\%$ & $22.18\%$ & $\textbf{8.28\%}$ & $1.81$\\

    6 & $0.87\%$ & $0.86\%$ &$21.96\%$ & $24.65\%$ & $9.55\%$ & $\textbf{0.86\%}$ & $1.00$\\

    8 & $22.61\%$ & $8.29\%$ &$84.54\%$ & $32.03\%$ & $38.20\%$ & $\textbf{23.61\%}$ & $2.85$\\
9 & $21.54\%$ & $5.03\%$ &$86.17\%$ & $12.50\%$ & $6.84\%$ & $\textbf{6.17\%}$ & $1.23$
  \end{tabular} }

\vspace{0.5em}

  \resizebox{\columnwidth}{!}{% <------ Don't forget this %
\centering
  \begin{tabular}{c|c|cccc|c}
    \hline
    Protected & Lower  & \multicolumn{4}{c|}{Upper Bounds} & Best \\
          Attribute      & Bound & Average & Greedy & Average$+$LM & Greedy$+$LM & Ratio \\
    \hline
    Attendant & $1.91\%$ & $14.59\%$ & $2.34\%$ & $2.10\%$ & $\textbf{2.08\%}$ & $1.09$\\ 
Father Race & $0.92\%$ & $19.75\%$ & $1.50\%$ & $1.32\%$ & $\textbf{1.28\%}$ & $1.40$\\ 
Mother Race & $0.65\%$ & $13.43\%$ & $1.31\%$ & $1.14\%$ & $\textbf{1.12\%}$ & $1.71$\\ 
Payer & $1.74\%$ & $24.15\%$ & $1.96\%$ & $1.97\%$ & $\textbf{1.89\%}$ & $1.09$\\ 
  \end{tabular}
}
\end{table}

Second, we used data about births in the United States \citep{CDCnatalitydata2017}, which provides detailed information about each birth that occurred during 2017. The data set includes approximately $3.8$ million data points and $50$ attributes. Using only attributes that are known before the labor, we generated three classifiers that attempt to predict the type of labor out of the five possible options (e.g., spontaneous, Cesarean): a decision tree (error $30.8\%$), a $k$-Nearest Neighbor classifier, where $k$ was set to $9$ using parameter tuning (error $24.6\%$), and a 2-layer neural network (error 22.43\%). This allows studying a high-error regime for the estimation of $\UNF$.
We estimated the $\UNF$ of each classifier, with respect to different protected attributes. 
\tabref{censusdec} (bottom), and \tabref{labornaive}, \tabref{labormlp} in \appref{tables}, report the \UNF\ lower bound and upper bounds relative to each of these protected attributes.
Here too, our proposed approach provides the tightest upper bounds. The ratio between the upper bound and the lower bound is between $1.00$ and $1.58$, showing that here too, \UNF\ is estimated to a high accuracy.

\begin{table}[t]
  \begin{center}
  \caption{\small Comparing the output of the $\minunf$ local optimizer (LO) to the $\UNF$ range calculated for US Census classifiers (top) and Natality (bottom). The ranges are derived from the results of \tabref{censusdec}, \tabref{censusnn} and \tabref{labornaive}.}
  \label{tab:censusdecmindisc}
 \resizebox{\columnwidth}{!}{% <------ Don't forget this %
  \begin{tabular}{cccccclc}
    \toprule
    US Census & \multicolumn{2}{c}{Decision tree} & \multicolumn{2}{c}{Nearest Neighbor}\\
    \cmidrule(lr){2-3} \cmidrule(lr){4-5}
    \# Labels &  $\minunf$ LO & true \UNF & $\minunf$ LO & true \UNF\\
    \midrule
    3 & $2.47\%$ & $5.39\%-7.65\%$        & $3.16\%$ & $6.56\%-9.77\%$\\
3 & $1.11\%$ & $4.35\%-5.28\%$         & $2.86\%$ &  $6.13\%-9.87\%$\\
3 &  $1.20\%$ & $3.24\%-3.25\%$         & $3.01\%$ & $6.82\%-9.59\%$\\
3 &  $0.84\%$ & $4.24\%-4.25\%$          & $2.79\%$ & $6.16\%-8.81\%$\\
3 &  $0.98\%$ & $2.65\%-3.81\%$            & $2.83\%$ & $7.15\%-9.05\%$\\
3 &  $1.13\%$ & $1.85\%$          & $2.82\%$ & $6.93\%-9.08\%$\\
3 &  $1.11\%$ & $1.96\%$          & $3.02\%$ & $6.56\%-8.59\%$\\
3 &  $0.68\%$ & $2.28\%$          &  $2.82\%$& $7.14\%-9.47\%$\\
3 &  $1.99\%$ & $4.57\%-6.10\%$            & $2.40\%$& $5.24\%-7.01\%$\\
4 & $1.07\%$ & $1.48\%-1.49\%$         & $4.29\%$ & $7.53\%-10.81\%$\\
5 &  $2.08\%$ & $2.06\%-3.91\%$           & $6.03\%$ &$8.86\%-10.74\%$\\
    5 & $3.13\%$ & $4.57\%-8.28\%$          & $3.29\%$ & $8.86\%-10.74\%$\\
    5 &  --   & -- &  $5.99\%$ & $9.50\%-18.22\%$\\
    5 &  --   & -- &        $5.62\%$ & $8.88\%-21.25\%$\\
    6 &  $0.32\%$ & $0.86\%$     & $6.41\%$ & $10.10\%-20.58\%$\\
8 & $4.08\%$ & $8.29\%-23.61\%$    &  $5.97\%$ & $7.89\%-20.45\%$\\
9 &  $2.38\%$ & $5.03\%-6.17\%$  &  $4.31\%$ & $7.91\%-14.09\%$\\
  \end{tabular}}
\end{center}
\begin{center}
    \resizebox{\columnwidth}{!}{% <------ Don't forget this %
  \begin{tabular}{ccccc}
    \toprule
    Natality & \multicolumn{2}{c}{Decision tree} & \multicolumn{2}{c}{$k$-Nearest-Neighbors}\\
     \cmidrule(lr){2-3} \cmidrule(lr){4-5}
    Protected~Attribute & $\minunf$ LO & true \UNF & $\minunf$ LO& true \UNF\\
    \hline
    Attendant & $0.75\%$ & $1.91\%-2.08\%$ & $0.74\%$ & $1.80\%-1.82\%$\\
Father Race & $0.26\%$ &  $0.92\%-1.28\%$ & $0.32\%$ & $1.17\%-1.18\%$\\
Mother Race & $0.12\%$ & $0.65\%-1.12\%$ & $0.14\%$ & $0.61\%-0.62\%$\\
    Payer & $0.05\%$ & $1.74\%-1.89\%$ & $0.59\%$ & $1.73\%-1.75\%$\\
  \end{tabular}}
  \end{center}
\end{table}

\subsection{Bounding \minunf\ without Confusion Matrices}\label{sec:multunknownexp}

In the second set of experiments, we used only the population-level frequencies $(\bw, \{(\bpi_a,\bhp_a)\}_{a \in \cA})$ to estimate $\minunf$. We ran the local optimization procedure provided in \secref{multunknown} to find an upper bound on $\minunf$, and compared this result to the lower bound \eqref{miny2}, to provide a range of possible values for $\minunf$. This range can indicate whether the population-level frequencies point to a possible, or definite, large deviation from multiclass equalized odds, as measured by the \UNF.

First, we calculated the local minimizer of $\minunf$ for the same two labeled data sets that were tested in \secref{multknownexp}, this time without access to the confusion matrices. We then compared the obtained value to the actual ranges of $\UNF$ that were calculated using the confusion matrices in \secref{multknownexp}. \tabref{censusdecmindisc}  provides results for the US Census data set and the Natality data set. In all but a single case
the local optimizer of $\minunf$ was lower than the true $\UNF$ range of the classifier, showing that this value is indeed a relevant best-case value for $\UNF$ with unknown confusion matrices.

\begin{figure}[h]
    \begin{center}
    \includegraphics[width=0.9\columnwidth]{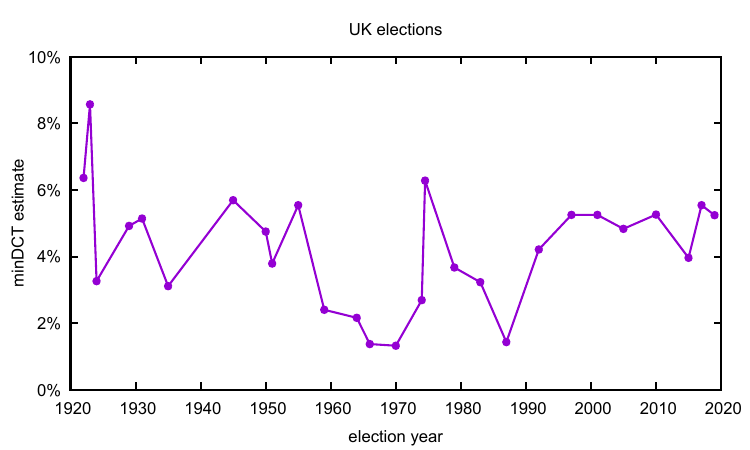}
    \caption{Calculated value $\minunf$ for each election year in the UK elections dataset, where the value is reported for the classifier that attempts to predict this year's election result using the results of the previous election.}
  \label{fig:ukgraph}
\end{center}
\end{figure}

Next, we report two experiments for which we do not have ground truth labels to compare to, to demonstrate how this method can be used for studying various empirical questions. 
In the first experiment, we used data about the general elections in the UK from 1918 to 2019 \citep{UKelectiondata}. We studied the changes in voting patterns between elections by
studying the $\minunf$ value of a hypothetical classifier that would predict the vote of individuals in one general election to be the same as their vote in the previous general elections (ignoring the change in population between elections). We studied $\UNF$ when the protected attribute was the geographic region of the voters, as reported in the data set.
A high \UNF\ value of such a classifier would indicate a possibly large difference between voting pattern changes across regions. \figref{ukgraph} shows the value of $\minunf$ by election year, revealing clear differences between different periods of the last century.\footnote{Note the spike in \figref{ukgraph} for Oct.~1974 elections; they were unusual as they were held in the same year as the previous elections, and resulted in a significant political change 
  \citep{RoeCrines2021}.} The full results are reported in \tabref{ukelect} in the appendix. In \appref{education}, we report an additional experiment, on a US education data set.

\section{Conclusion}\label{sec:discussion}
In this work, we provided a definition and bounding methods for the \UNF\ measure for multiclass classifiers. This provides a new tool for auditing fairness in multiclass classification. Because \UNF\ is interpretable as a fraction of the population, the estimation methods that we proposed can be used to provide a clear evaluation of classifiers' deviation from equalized odds, even when there are many classes, protected attribute values, or highly imbalanced data sets.

\section*{Impact Statement}

This paper studies the auditing of classifiers for fairness in the sense of multiclass equalized odds. Identifying classifiers that may violate fairness criteria is an important task that can help advance societal desiderata. Nonetheless, such use must always consider the suitability of the fairness criterion to the specific classification task. In addition, in some cases the methods that we propose may provide loose bounds for the studied criterion. These cases can be identified by observing a large gap between the upper bound and the lower bound. In such cases, care must be taken when making operative decisions based on the results of the methods.

\section*{Acknowledgements}

We acknowledge the support of the Natural Sciences and Engineering Research Council of Canada (NSERC), [funding reference number RGPIN-2024-05907]. Resources used in preparing this research were provided, in part, by the Province of Ontario, the Government of Canada through CIFAR, and companies sponsoring the Vector Institute; see \url{https://vectorinstitute.ai/partnerships/current-partners/}. This work was supported by the Israeli Council for Higher Education (CHE) via the Data Science Research Center at Ben-Gurion University.

\bibliography{mybib}

\begin{thebibliography}{40}
\providecommand{\natexlab}[1]{#1}
\providecommand{\url}[1]{\texttt{#1}}
\expandafter\ifx\csname urlstyle\endcsname\relax
  \providecommand{\doi}[1]{doi: #1}\else
  \providecommand{\doi}{doi: \begingroup \urlstyle{rm}\Url}\fi

\bibitem[Alghamdi et~al.(2022)Alghamdi, Hsu, Jeong, Wang, Michalak, Asoodeh,
  and Calmon]{AlghamadiEtAl22}
Alghamdi, W., Hsu, H., Jeong, H., Wang, H., Michalak, P., Asoodeh, S., and
  Calmon, F.
\newblock Beyond adult and compas: Fair multi-class prediction via information
  projection.
\newblock In Koyejo, S., Mohamed, S., Agarwal, A., Belgrave, D., Cho, K., and
  Oh, A. (eds.), \emph{Advances in Neural Information Processing Systems},
  volume~35, pp.\  38747--38760. Curran Associates, Inc., 2022.

\bibitem[Angwin et~al.(2022)Angwin, Larson, Mattu, and
  Kirchner]{angwin2022machine}
Angwin, J., Larson, J., Mattu, S., and Kirchner, L.
\newblock Machine bias.
\newblock In \emph{Ethics of data and analytics}, pp.\  254--264. Auerbach
  Publications, 2022.

\bibitem[Awasthi et~al.(2021)Awasthi, Beutel, Kleindessner, Morgenstern, and
  Wang]{awasthi2021evaluating}
Awasthi, P., Beutel, A., Kleindessner, M., Morgenstern, J., and Wang, X.
\newblock Evaluating fairness of machine learning models under uncertain and
  incomplete information.
\newblock In \emph{Proceedings of the 2021 ACM conference on fairness,
  accountability, and transparency}, pp.\  206--214, 2021.

\bibitem[Bellamy et~al.(2019)Bellamy, Dey, Hind, Hoffman, Houde, Kannan, Lohia,
  Martino, Mehta, Mojsilovi{\'c}, et~al.]{Bellamy19}
Bellamy, R.~K., Dey, K., Hind, M., Hoffman, S.~C., Houde, S., Kannan, K.,
  Lohia, P., Martino, J., Mehta, S., Mojsilovi{\'c}, A., et~al.
\newblock {AI} fairness 360: An extensible toolkit for detecting and mitigating
  algorithmic bias.
\newblock \emph{IBM Journal of Research and Development}, 63\penalty0
  (4/5):\penalty0 4--1, 2019.

\bibitem[Black et~al.(2022)Black, Elzayn, Chouldechova, Goldin, and
  Ho]{BlackElChGoHo22}
Black, E., Elzayn, H., Chouldechova, A., Goldin, J., and Ho, D.
\newblock Algorithmic fairness and vertical equity: Income fairness with irs
  tax audit models.
\newblock In \emph{Proceedings of the 2022 ACM Conference on Fairness,
  Accountability, and Transparency}, FAccT '22, pp.\  1479–1503, New York,
  NY, USA, 2022. Association for Computing Machinery.
\newblock ISBN 9781450393522.

\bibitem[Calmon et~al.(2017)Calmon, Wei, Vinzamuri, Natesan~Ramamurthy, and
  Varshney]{Calmonetal2017}
Calmon, F., Wei, D., Vinzamuri, B., Natesan~Ramamurthy, K., and Varshney, K.~R.
\newblock Optimized pre-processing for discrimination prevention.
\newblock In Guyon, I., Luxburg, U.~V., Bengio, S., Wallach, H., Fergus, R.,
  Vishwanathan, S., and Garnett, R. (eds.), \emph{Advances in Neural
  Information Processing Systems 30}, pp.\  3992--4001. Curran Associates,
  Inc., 2017.

\bibitem[Caton \& Haas(2024)Caton and Haas]{caton2024fairness}
Caton, S. and Haas, C.
\newblock Fairness in machine learning: A survey.
\newblock \emph{ACM Computing Surveys}, 56\penalty0 (7):\penalty0 1--38, 2024.

\bibitem[{CDC}(2017)]{CDCnatalitydata2017}
{CDC}.
\newblock United states natality public use file, 2017.
\newblock URL
  \url{https://ftp.cdc.gov/pub/Health_Statistics/NCHS/Datasets/DVS/natality/Nat2017us.zip}.

\bibitem[Charalambous \& Conn(1978)Charalambous and
  Conn]{charalambous1978efficient}
Charalambous, C. and Conn, A.
\newblock An efficient method to solve the minimax problem directly.
\newblock \emph{SIAM Journal on Numerical Analysis}, 15\penalty0 (1):\penalty0
  162--187, 1978.

\bibitem[Cherian \& Cand{\`e}s(2024)Cherian and
  Cand{\`e}s]{cherian2024statistical}
Cherian, J.~J. and Cand{\`e}s, E.~J.
\newblock Statistical inference for fairness auditing.
\newblock \emph{Journal of Machine Learning Research}, 25\penalty0
  (149):\penalty0 1--49, 2024.

\bibitem[Cornacchia et~al.(2023)Cornacchia, Anelli, Biancofiore, Narducci,
  Pomo, Ragone, and Di~Sciascio]{cornacchia2023auditing}
Cornacchia, G., Anelli, V.~W., Biancofiore, G.~M., Narducci, F., Pomo, C.,
  Ragone, A., and Di~Sciascio, E.
\newblock Auditing fairness under unawareness through counterfactual reasoning.
\newblock \emph{Information Processing \& Management}, 60\penalty0
  (2):\penalty0 103224, 2023.

\bibitem[Dastin(2022)]{dastin2022amazon}
Dastin, J.
\newblock Amazon scraps secret ai recruiting tool that showed bias against
  women.
\newblock In \emph{Ethics of data and analytics}, pp.\  296--299. Auerbach
  Publications, 2022.

\bibitem[Denis et~al.(2024)Denis, Elie, Hebiri, and Hu]{denis2024fairness}
Denis, C., Elie, R., Hebiri, M., and Hu, F.
\newblock Fairness guarantees in multi-class classification with demographic
  parity.
\newblock \emph{Journal of Machine Learning Research}, 25\penalty0
  (130):\penalty0 1--46, 2024.

\bibitem[Donini et~al.(2018)Donini, Oneto, Ben-David, Shawe-Taylor, and
  Pontil]{DoniniOnBeShPo18}
Donini, M., Oneto, L., Ben-David, S., Shawe-Taylor, J.~S., and Pontil, M.
\newblock Empirical risk minimization under fairness constraints.
\newblock In Bengio, S., Wallach, H., Larochelle, H., Grauman, K.,
  Cesa-Bianchi, N., and Garnett, R. (eds.), \emph{Advances in Neural
  Information Processing Systems 31}, pp.\  2791--2801. Curran Associates,
  Inc., 2018.

\bibitem[Dua \& Graff(2019)Dua and Graff]{UCI2019}
Dua, D. and Graff, C.
\newblock {UCI} machine learning repository, 2019.
\newblock URL \url{http://archive.ics.uci.edu/ml}.

\bibitem[Fabris et~al.(2023)Fabris, Esuli, Moreo, and
  Sebastiani]{fabris2023measuring}
Fabris, A., Esuli, A., Moreo, A., and Sebastiani, F.
\newblock Measuring fairness under unawareness of sensitive attributes: A
  quantification-based approach.
\newblock \emph{Journal of Artificial Intelligence Research}, 76:\penalty0
  1117--1180, 2023.

\bibitem[Goyal et~al.(2022)Goyal, Soriano, Hazirbas, Sagun, and
  Usunier]{GoyalSoHaSaUs22}
Goyal, P., Soriano, A.~R., Hazirbas, C., Sagun, L., and Usunier, N.
\newblock Fairness indicators for systematic assessments of visual feature
  extractors.
\newblock In \emph{Proceedings of the 2022 ACM Conference on Fairness,
  Accountability, and Transparency}, FAccT '22, pp.\  70–88, New York, NY,
  USA, 2022. Association for Computing Machinery.
\newblock ISBN 9781450393522.

\bibitem[Grandinetti(2023)]{grandinetti2023examining}
Grandinetti, J.
\newblock Examining embedded apparatuses of ai in facebook and tiktok.
\newblock \emph{{AI} \& Society}, pp.\  1--14, 2023.

\bibitem[Hall et~al.(2023)Hall, Galabova, Gottwald, and
  Feldmeier]{hall2023highs}
Hall, J., Galabova, I., Gottwald, L., and Feldmeier, M.
\newblock Highs--high performance software for linear optimization, 2023.

\bibitem[Hardt et~al.(2016)Hardt, Price, and Srebro]{HardtPrSr16}
Hardt, M., Price, E., and Srebro, N.
\newblock Equality of opportunity in supervised learning.
\newblock In \emph{Advances in neural information processing systems}, pp.\
  3315--3323, 2016.

\bibitem[Jung et~al.(2021)Jung, Lee, Park, and Moon]{jung2021fair}
Jung, S., Lee, D., Park, T., and Moon, T.
\newblock Fair feature distillation for visual recognition.
\newblock In \emph{Proceedings of the IEEE/CVF conference on computer vision
  and pattern recognition}, pp.\  12115--12124, 2021.

\bibitem[Kafuria(2022)]{kafuria2022predictice}
Kafuria, A.~D.
\newblock \emph{A predictice model for health insurance premium rates using
  machine learning algorithms}.
\newblock PhD thesis, University of Rwanda, 2022.

\bibitem[Kiessling et~al.(2022)Kiessling, Zanelli, Nurkanovi{\'c}, Gillis,
  Diehl, Zeilinger, Pipeleers, and Swevers]{kiessling2022feasible}
Kiessling, D., Zanelli, A., Nurkanovi{\'c}, A., Gillis, J., Diehl, M.,
  Zeilinger, M., Pipeleers, G., and Swevers, J.
\newblock A feasible sequential linear programming algorithm with application
  to time-optimal path planning problems.
\newblock In \emph{2022 IEEE 61st Conference on Decision and Control (CDC)},
  pp.\  1196--1203. IEEE, 2022.

\bibitem[Kleinberg et~al.(2017)Kleinberg, Mullainathan, and
  Raghavan]{KleinbergMuRa17}
Kleinberg, J., Mullainathan, S., and Raghavan, M.
\newblock Inherent trade-offs in the fair determination of risk scores.
\newblock In \emph{8th Innovations in Theoretical Computer Science Conference
  (ITCS 2017)}. Schloss Dagstuhl-Leibniz-Zentrum fuer Informatik, 2017.

\bibitem[Menon \& Williamson(2018)Menon and Williamson]{MenonWi18}
Menon, A.~K. and Williamson, R.~C.
\newblock The cost of fairness in binary classification.
\newblock In \emph{Conference on Fairness, Accountability and Transparency},
  pp.\  107--118, 2018.

\bibitem[Nocedal \& Wright(2006)Nocedal and Wright]{nocedal2006numerical}
Nocedal, J. and Wright, S.
\newblock \emph{Numerical optimization}.
\newblock Springer Science \& Business Media, 2006.

\bibitem[Pinz{\'o}n et~al.(2024)Pinz{\'o}n, Palamidessi, Piantanida, and
  Valencia]{pinzon2024incompatibility}
Pinz{\'o}n, C., Palamidessi, C., Piantanida, P., and Valencia, F.
\newblock On the incompatibility of accuracy and equal opportunity.
\newblock \emph{Machine Learning}, 113\penalty0 (5):\penalty0 2405--2434, 2024.

\bibitem[Pleiss et~al.(2017)Pleiss, Raghavan, Wu, Kleinberg, and
  Weinberger]{PleissRaKlKi17}
Pleiss, G., Raghavan, M., Wu, F., Kleinberg, J., and Weinberger, K.~Q.
\newblock On fairness and calibration.
\newblock In \emph{Advances in Neural Information Processing Systems}, pp.\
  5680--5689, 2017.

\bibitem[Putzel \& Lee(2022)Putzel and Lee]{putzel2022blackbox}
Putzel, P. and Lee, S.
\newblock Blackbox post-processing for multiclass fairness.
\newblock \emph{arXiv preprint arXiv:2201.04461}, 2022.

\bibitem[Roe-Crines(2021)]{RoeCrines2021}
Roe-Crines, A.~S.
\newblock \emph{Who Governs? The General Election Defeats of 1974}, pp.\
  355--375.
\newblock Springer International Publishing, Cham, 2021.

\bibitem[Roth et~al.(2022)Roth, Saint-Jacques, and Yu]{RothSaYu22}
Roth, J., Saint-Jacques, G., and Yu, Y.
\newblock An outcome test of discrimination for ranked lists.
\newblock In \emph{Proceedings of the 2022 ACM Conference on Fairness,
  Accountability, and Transparency}, FAccT '22, pp.\  350–356, New York, NY,
  USA, 2022. Association for Computing Machinery.
\newblock ISBN 9781450393522.

\bibitem[Rouzot et~al.(2022)Rouzot, Ferry, and Huguet]{rouzot2022learning}
Rouzot, J., Ferry, J., and Huguet, M.-J.
\newblock Learning optimal fair scoring systems for multi-class classification.
\newblock In \emph{2022 IEEE 34th International Conference on Tools with
  Artificial Intelligence (ICTAI)}, pp.\  197--204. IEEE, 2022.

\bibitem[Sabato \& Yom-Tov(2020)Sabato and Yom-Tov]{SabatoYo20}
Sabato, S. and Yom-Tov, E.
\newblock Bounding the fairness and accuracy of classifiers from population
  statistics.
\newblock In III, H.~D. and Singh, A. (eds.), \emph{Proceedings of the 37th
  International Conference on Machine Learning}, volume 119 of
  \emph{Proceedings of Machine Learning Research}, pp.\  8316--8325. PMLR,
  13--18 Jul 2020.

\bibitem[Saleiro et~al.(2018)Saleiro, Kuester, Hinkson, London, Stevens,
  Anisfeld, Rodolfa, and Ghani]{saleiro2018aequitas}
Saleiro, P., Kuester, B., Hinkson, L., London, J., Stevens, A., Anisfeld, A.,
  Rodolfa, K.~T., and Ghani, R.
\newblock Aequitas: A bias and fairness audit toolkit.
\newblock \emph{arXiv preprint arXiv:1811.05577}, 2018.

\bibitem[Taskesen et~al.(2021)Taskesen, Blanchet, Kuhn, and
  Nguyen]{taskesen2021statistical}
Taskesen, B., Blanchet, J., Kuhn, D., and Nguyen, V.~A.
\newblock A statistical test for probabilistic fairness.
\newblock In \emph{Proceedings of the 2021 ACM conference on fairness,
  accountability, and transparency}, pp.\  648--665, 2021.

\bibitem[{USDA Economic Research Service}(2021)]{USeducationdata}
{USDA Economic Research Service}.
\newblock Highest level of educational attainment, 2021.
\newblock URL \url{https://data.ers.usda.gov/reports.aspx?ID=17829}.

\bibitem[Wang et~al.(2024)Wang, He, Gao, and Calmon]{wang2024aleatoric}
Wang, H., He, L., Gao, R., and Calmon, F.
\newblock Aleatoric and epistemic discrimination: Fundamental limits of
  fairness interventions.
\newblock \emph{Advances in Neural Information Processing Systems}, 36, 2024.

\bibitem[Wang et~al.(2021)Wang, Liu, and Levy]{wang2021fair}
Wang, J., Liu, Y., and Levy, C.
\newblock Fair classification with group-dependent label noise.
\newblock In \emph{Proceedings of the 2021 ACM conference on fairness,
  accountability, and transparency}, pp.\  526--536, 2021.

\bibitem[Watson et~al.(2020)Watson, Uberoi, and Loft]{UKelectiondata}
Watson, C., Uberoi, E., and Loft, P.
\newblock General election results from 1918 to 2019, 2020.
\newblock URL
  \url{https://commonslibrary.parliament.uk/research-briefings/cbp-8647/}.

\bibitem[Xue(2023)]{xue2023group}
Xue, Z.
\newblock Group adaboost with fairness constraint.
\newblock In \emph{Proceedings of the 2023 SIAM International Conference on
  Data Mining (SDM)}, pp.\  865--873. SIAM, 2023.

\end{thebibliography}
\bibliographystyle{icml2025}

\newpage
\appendix
\onecolumn

\section{Deferred Experiment Results}\label{app:tables}

\tabref{censusnn}---\tabref{ukelect} below provide the remaining result of the experiments which are described in \secref{exps}.

\begin{table}[h]
    \caption{Bounding the \UNF\ for multiclass classifiers with known confusion matrices; US Census data set with nearest neighbor classifiers. Table columns as in \tabref{censusdec}.}
  \label{tab:censusnn}
  
  \centering
  \begin{tabular}{c|c|c|cccc|c}
    \hline
    \#      & Error & Lower  & \multicolumn{4}{c|}{Upper Bounds} & Best \\
    Labels  &       & Bound  & Average &  Greedy & Average$+$LM &  Greedy$+$LM &  Ratio \\
    \hline
3 & $27.71\%$ & $6.56\%$ &$19.71\%$ & $10.64\%$ & $10.80\%$ & $\textbf{9.77\%}$ & $1.49$\\
3 & $28.71\%$ & $6.13\%$ &$19.00\%$ & $11.66\%$ & $10.67\%$ & $\textbf{9.87\%}$ & $1.61$\\
3 & $26.22\%$ & $6.82\%$ &$18.44\%$ & $10.46\%$ & $11.82\%$ & $\textbf{9.59\%}$ & $1.41$\\
3 & $24.61\%$ & $6.16\%$ &$19.54\%$ & $10.05\%$ & $11.23\%$ & $\textbf{8.81\%}$ & $1.43$\\
3 & $24.24\%$ & $7.15\%$ &$20.04\%$ & $10.44\%$ & $12.04\%$ & $\textbf{9.05\%}$ & $1.27$\\
3 & $23.52\%$ & $6.93\%$ &$20.42\%$ & $10.00\%$ & $11.97\%$ & $\textbf{9.08\%}$ & $1.31$\\
3 & $23.43\%$ & $6.56\%$ &$19.15\%$ & $8.91\%$ & $11.10\%$ & $\textbf{8.59\%}$ & $1.31$\\
3 & $23.46\%$ & $7.14\%$ &$20.01\%$ & $10.95\%$ & $12.08\%$ & $\textbf{9.47\%}$ & $1.33$\\
    3 & $19.45\%$ & $5.24\%$ &$23.24\%$ & $8.80\%$ & $9.98\%$ & $\textbf{7.01\%}$ & $1.34$\\
    4 & $27.22\%$ & $5.61\%$ &$39.16\%$ & $11.24\%$ & $18.47\%$ & $\textbf{9.12\%}$ & $1.63$\\
5 & $13.87\%$ & $7.53\%$ &$64.62\%$ & $13.32\%$ & $14.62\%$ & $\textbf{10.81\%}$ & $1.43$\\
5 & $17.54\%$ & $8.86\%$ &$59.21\%$ & $11.83\%$ & $29.39\%$ & $\textbf{10.74\%}$ & $1.21$\\
5 & $48.57\%$ & $9.50\%$ &$41.93\%$ & $22.14\%$ & $28.16\%$ & $\textbf{18.22\%}$ & $1.92$\\
    5 & $54.92\%$ & $8.88\%$ &$36.77\%$ & $24.23\%$ & $24.88\%$ & $\textbf{21.25\%}$ & $2.39$\\
    6 & $50.47\%$ & $10.10\%$ &$48.96\%$ & $25.51\%$ & $33.19\%$ & $\textbf{20.58\%}$ & $2.04$\\
8 & $42.67\%$ & $7.89\%$ &$69.92\%$ & $30.60\%$ & $47.78\%$ & $\textbf{20.45\%}$ & $2.59$\\
9 & $39.56\%$ & $7.91\%$ &$99.27\%$ & $16.90\%$ & $17.99\%$ & $\textbf{14.09\%}$ & $1.78$
  \end{tabular}
\end{table}

\begin{table}[h]
   \begin{center}
       \caption{Bounding the \UNF\ for multiclass classifiers with known confusion matrices; Natality data set with a $k$-Nearest-Neighbor classifier. Table columns as in \tabref{censusdec}.}
  \label{tab:labornaive}
  
  \centering
  \begin{tabular}{c|c|cccc|c}
    \hline
    Test & Lower  & \multicolumn{4}{c|}{Upper Bounds} & Best\\
         & Bound & Average & Greedy & Average$+$LM &  Greedy$+$LM & Ratio \\
    \hline
    Attendant & $1.80\%$ &$28.53\%$ & $\textbf{1.82\%}$ & $\textbf{1.82\%}$ & $\textbf{1.82\%}$ & $1.01$\\
Father Race & $1.17\%$ &$46.54\%$ & $1.20\%$ & $\textbf{1.18\%}$ & $\textbf{1.18\%}$ & $1.01$\\
Mother Race & $0.61\%$ &$21.47\%$ & $0.69\%$ & $\textbf{0.62\%}$ & $\textbf{0.62\%}$ & $1.02$\\
Payer & $1.73\%$ &$54.61\%$ & $1.79\%$ & $1.76\%$ & $\textbf{1.75\%}$ & $1.01$
  \end{tabular}
  %}
\end{center}
\end{table}

\begin{table}[h]
  \begin{center}
  \caption{Bounding the \UNF\ for multiclass classifiers with known confusion matrices; US Census data set with the neural network-based classifiers.  Table columns as in \tabref{censusdec}.}
  \label{tab:censusmlp}
\begin{tabular}{c|c|c|cccc|c}
    \hline
    \#     & Error & Lower & \multicolumn{4}{c|}{Upper Bounds} & Best Ratio \\
    Labels &       & Bound & Average & Greedy & Average+LM & Greedy+LM &    \\
    \hline
    3 & $12.76\%$ & $2.36\%$ &$20.19\%$ & $3.46\%$ & $3.04\%$ & $\textbf{3.03\%}$ & $1.28$\\
    3 & $12.12\%$ & $1.43\%$ &$23.16\%$ & $2.06\%$ & $2.75\%$ & $\textbf{1.84\%}$ & $1.28$\\
    3 & $8.95\%$ & $1.27\%$ &$18.89\%$ & $2.05\%$ & $1.75\%$ & $\textbf{1.75\%}$ & $1.38$\\
    3 & $8.96\%$ & $1.52\%$ &$16.63\%$ & $1.94\%$ & $1.83\%$ & $\textbf{1.77\%}$ & $1.16$\\
    3 & $6.04\%$ & $2.42\%$ &$22.88\%$ & $3.52\%$ & $\textbf{2.96\%}$ & $\textbf{2.96\%}$ & $1.22$\\
    3 & $8.26\%$ & $0.86\%$ &$34.80\%$ & $3.53\%$ & $\textbf{2.24\%}$ & $\textbf{2.24\%}$ & $2.60$\\
    3 & $13.31\%$ & $0.01\%$ &$68.05\%$ & $27.12\%$ & $\textbf{0.01\%}$ & $0.02\%$ & $2.45$\\
    3 & $8.07\%$ & $0.63\%$ &$24.58\%$ & $1.41\%$ & $0.79\%$ & $\textbf{0.78\%}$ & $1.23$\\
    3 & $10.39\%$ & $1.53\%$ &$40.93\%$ & $1.56\%$ & $2.06\%$ & $\textbf{1.56\%}$ & $1.02$ \\
    4 & $4.08\%$ & $0.93\%$ &$41.29\%$ & $4.07\%$ & $2.51\%$ & $\textbf{2.17\%}$ & $2.33$ \\
    5 & $4.09\%$ & $0.78\%$ &$69.65\%$ & $12.97\%$ & $1.89\%$ & $\textbf{0.99\%}$ & $1.28$\\
    5 & $5.40\%$ & $2.44\%$ &$66.73\%$ & $41.57\%$ & $3.18\%$ & $\textbf{2.53\%}$ & $1.04$\\
    5 & $17.48\%$ & $3.55\%$ &$46.89\%$ & $4.35\%$ & $4.24\%$ & $\textbf{4.05\%}$ & $1.14$\\
    5 & $21.35\%$ & $6.02\%$ &$64.82\%$ & $8.01\%$ & $8.86\%$ & $\textbf{7.67\%}$ & $1.27$ \\
    6 & $17.76\%$ & $2.49\%$ &$66.61\%$ & $6.01\%$ & $4.30\%$ & $\textbf{3.67\%}$ & $1.47$ \\
    8 & $26.69\%$ & $2.64\%$ &$68.10\%$ & $14.22\%$ & $5.66\%$ & $\textbf{4.28\%}$ & $1.62$ \\
    9 & $18.19\%$ & $2.95\%$ &$80.76\%$ & $3.13\%$ & $4.15\%$ & $\textbf{3.12\%}$ & $1.06$
\end{tabular}
\end{center}
\end{table}

\begin{table}[h]
  \begin{center}
  \caption{Bounding the \UNF\ for multiclass classifiers with known confusion matrices; Natality data set with the neural network classifier. Table columns as in \tabref{censusdec}.}
  \label{tab:labormlp}
\begin{tabular}{c|c|cccc|c}
    \hline
    Protected  & Lower & \multicolumn{4}{c|}{Upper Bounds}         & Best \\
    Attribute  & Bound & Average & Greedy & Average+LM & Greedy+LM & Ratio   \\
    \hline
    Attendant   & $1.03\%$ &$26.25\%$ & $1.05\%$ & $\textbf{1.04\%}$ & $\textbf{1.04\%}$ & $1.01$\\
    Father Race & $0.67\%$ &$50.53\%$ & $0.73\%$ & $\textbf{0.67\%}$ & $\textbf{0.67\%}$ & $1.01$\\
    Mother Race & $0.58\%$ &$13.25\%$ & $0.58\%$ & $\textbf{0.58\%}$ & $\textbf{0.58\%}$ & $1.01$\\
    Payer       & $1.21\%$ &$55.86\%$ & $1.21\%$ & $\textbf{1.21\%}$ & $\textbf{1.21\%}$ & $1.00$ \\
\end{tabular}
\end{center}
\end{table}

\begin{table}[h]
  \caption{Comparing the output of the minDCP local optimizer (LO) to the DCP range calculated for US Census classifiers for the neural network classifiers. The ranges are derived from \tabref{censusmlp}. }
\centering
\begin{tabular}{c c c}
\hline
US Census & \multicolumn{2}{c}{Neural Network} \\
\cline{2-3}
\# Labels & minDCP LO & true DCP \\
\hline
3 & 0.97\% & 2.36\% -- 3.03\% \\
3 & 0.63\% & 1.43\% -- 1.84\% \\
3 & 0.68\% & 1.27\% -- 1.75\% \\
3 & 0.53\% & 1.52\% -- 1.77\% \\
3 & 1.23\% & 2.42\% -- 2.96\% \\
3 & 0.56\% & 0.86\% -- 2.24\% \\
3 & 0.01\% & 0.01\% \\
3 & 0.26\% & 0.63\% -- 0.78\% \\
3 & 0.90\% & 1.53\% -- 1.56\% \\
4 & 0.47\% & 0.93\% -- 2.17\% \\
5 & 0.70\% & 0.78\% -- 0.99\% \\
5 & 1.36\% & 2.44\% -- 2.53\% \\
5 & 1.09\% & 3.55\% -- 4.05\% \\
5 & 3.16\% & 6.02\% -- 7.67\% \\
6 & 0.80\% & 2.49\% -- 3.67\% \\
8 & 1.77\% & 2.64\% -- 4.28\% \\
9 & 1.63\% & 2.95\% -- 3.12\% \\
\end{tabular}
\end{table}

\begin{table}[h]
  \caption{Comparing the output of the minDCP local optimizer (LO) to the DCP range calculated for the Natality data set for the neural network classifier. The ranges are derived from \tabref{labormlp}.}
  \centering
\begin{tabular}{c c c}
\hline
Natality & \multicolumn{2}{c}{Neural Network} \\
\cline{2-3}
Protected Attribute & minDCP LO & true DCP \\
\hline
Attendant           & 0.15\% & 1.03\% -- 1.04\% \\
Father Race         & 0.24\% & 0.67\% \\
Mother Race         & 0.10\% & 0.58\% \\
Payer               & 0.13\% & 1.21\% \\
\end{tabular}
\end{table}

  \begin{table}[h]
  \begin{center}
    \caption{Calculated $\minunf$ local optimizers for the UK elections data. In each line, the election data from the listed baseline year was used to predict the vote in the listed prediction year.}
  \label{tab:ukelect}
   
  \begin{tabular}{llccc}
    \multicolumn{2}{c}{Election years} & & \\
    \cmidrule(lr){1-2}
    Baseline & Prediction & \# protected attribute values & $\minunf$ LO\\
     1918  &  1922  & 12 & $6.36\%$\\
 1922  &  1923  & 13& $8.57\%$\\
 1923  &  1924  & 13& $3.26\%$\\
 1924  &  1929  & 13& $4.92\%$\\
 1929  &  1931  & 13& $5.14\%$\\
 1931  &  1935  & 13& $3.11\%$\\
 1935  &  1945  & 13& $5.69\%$\\
 1945  &  1950  & 12& $4.75\%$\\
 1950  &  1951  & 12& $3.79\%$\\
 1951  &  1955  & 12& $5.54\%$\\
  1955  &  1959  & 12& $2.40\%$\\
  1959  &  1964  & 12& $2.16\%$\\
  1964  &  1966  & 12& $1.37\%$\\
  1966  &  1970  & 12& $1.32\%$\\
  1970  &  1974 (Feb)  & 12& $2.69\%$\\
  1974  (Feb) &  1974 (Oct)  & 12& $6.28\%$\\
  1974 (Oct)  &  1979  & 12& $3.67\%$\\
  1979  &  1983  & 11& $3.23\%$\\
  1983  &  1987  & 12& $1.43\%$\\
  1987  &  1992  & 12& $4.21\%$\\
  1992  &  1997  & 11& $5.25\%$\\
  1997  &  2001  & 12& $5.25\%$\\
  2001  &  2005  & 12& $4.83\%$\\
  2005  &  2010  & 12& $5.26\%$\\
  2010  &  2015  & 12& $3.96\%$\\
  2015  &  2017  & 12& $5.54\%$\\
  2017  &  2019  & 11& $5.24\%$\\
  \midrule
  \end{tabular}
  
\end{center}
\end{table}

\clearpage
\section{An Additional Experiment}\label{app:education}

In this experiment, we studied a data set on US education \citep{USeducationdata}, which provides the percentage of various levels of education attainment (e.g., high school, college) in each US state in each decade. Here too, we calculated $\minunf$ for a hypothesized classifier that predicts the education level to be distributed the same in each state in each decade. The protected attribute as set to be the state. \tabref{education} provides our results. Here, we found no significant differences in the $\UNF$ of change patterns in different decades, indicating a fairly constant behavior of this measure of divergence between states. This type of analysis can be used for exploratory research on social questions.

\begin{table}[H]
  \caption{Calculated $\minunf$ upper bounds for the US education data set.}
  \label{tab:education}
  \begin{center}
  \begin{tabular}{llc}
    \toprule
    \multicolumn{2}{c}{Year} &  \\
    \cmidrule(lr){1-2}
    Baseline & Predicted &  $\minunf$ upper bound\\
    1970 & 1980 & $2.38\%$\\
    1980 & 1990 &$2.94\%$\\
    1990 & 2000 &$2.22\%$\\
    2000 & 2015-2019 & $2.32\%$\\
    \midrule
  \end{tabular}
\end{center}
\end{table}

\section{The Greedy Initialization Procedure} \label{app:greedy}

We provide here the full details of the greedy initialization procedure presented in \secref{multiclassknown}.

Let $\bof := \{f_{y}\}_{y \in \cY}$, where $f_{y}: \cY \rightarrow \cY$, be label mappings conditioned on the true label $y$, which can
map some predicted labels to the same transformed label.
For a given classifier $\cC$ with distribution $\cD$, let $\cC[\bof]$ be a hypothetical classifier that predicts $f_y(\hat{y})$ whenever the true label is $y$ and $\cC$ would predict $\hat{y}$. For a given distribution $\cP$ over $(Y, \hat{Y}, A)$, let $\cP[\bof]$ be the distribution of $(Y,f_{Y}(\hat{Y}),A)$. Then, $\cD[\bof]$ is the distribution determined by $\cC[\bof]$. 
It is easy to see that  $\UNF(\cC[\bof]) \leq \UNF(\cC).$ This is because the equality $\cD_a^y = (1-\eta^y_a)\dbase^y  + \eta_a^y \cN^y_a$ implies that also $\cD_a^y[\bof] = (1-\eta^y_a)\dbase^y[\bof]  + \eta_a^y \cN^y_a[\bof]$. Thus, minimizing over $\eta_a^y$ for $\cC[\bof]$ can never result in a solution of a higher value than minimizing for $\cC$. We use this observation to devise an iterative greedy optimization procedure.

For $i \in [k-1]$, let $\bof_i := (f_{i,y})_{y \in \cY}$ be an indexed set of label mappings, $f_{i,y}: \cY \rightarrow \cY$, defined as follows. Let $y_i$ be the $i$'th label in $\cY$ that is different from $y$. Denote $\cY_i = \{y,y_1,\ldots,y_i\}$. Note that $\cY_{k-1}=\cY$. For $i \in [k-2]$, define 
$f_{i,y}(j) = j\cdot \one[j \in \cY_i] + y_i \cdot \one[j \notin \cY_i].$
Note that $f_{i,y}$ can be calculated from the image of $f_{i-1,y}$. Hence, $\cC[\bof_i]$ is a refinement of $\cC[\bof_{i-1}]$. In addition, $f_{k-1,y}$ is the identity.
Thus, the following monotonicity property holds:
\begin{align*}
  &\UNF(\cC[\bof_1]) \leq \UNF(\cC[\bof_2]) \leq \ldots \leq \UNF(\cC[\bof_{k-1}]) = \UNF(\cC).
\end{align*}
Moreover, $\UNF(\bof_1)$ can be calculated exactly as in case of binary classification, since the range of $\bof_1$ includes only $y$ and $y_1$.
Based on these observations, we define a greedy procedure for calculating an assignment for $\alla^y$ to initialize the minimization in \eqref{unfairmin}.

Let $\allabigg^y[i]$ be row $y$ of the confusion matrices of $\cD[\bof_i]$. Then coordinates $j \in \cY_{i-1}$ of $\alla^y[i]$ are the same as those of $\allabigg^y$, and coordinate $y_i$ has the value $\widetilde{\alpha}_a^{yy_{i}}:=\sum_{j=i}^{k-1} \alpha_a^{yy_j}$. We have
\begin{align*}
  &\UNF_y(\allabigg^y[1]) \leq \UNF_y(\allabigg^y[2]) \leq \ldots\leq \UNF_y(\allabigg^y[k-1]) =\UNF_y(\allabigg^y).
\end{align*}

  The greedy procedure first calculates an assignment for $\alla^y[1]$ that obtains the value of $\UNF_y(\allabigg^y[1])$. This is a binary problem, which can be solved exactly following \citet{SabatoYo20}. Then, at each iteration $i+1$ for $i \in [k-2]$, a local minimum $\allag^y[i+1]$ for \mbox{$\UNF_y(\allabigg^y[i+1])$} is calculated by constraining $\alla^y[i+1]$ to have the same coordinates as $\alla^y[i]$ on $\cY_{i-1}$, and minimizing over $\alphabase^{yy_i},\alphabase^{yy_{i+1}}$ such that their sum is equal to coordinate $y_{i}$ in $\alla^y[i]$. This minimization can be solved exactly, as follows.

Denote the value of coordinate $y_i$ in $\alla^y[i]$ by $\gamma = 1-\sum_{j \in \cY_{i-1}}\alphabase^{yj}$. Minimizing the objective of \mbox{$\UNF_y(\allabigg^y[i+1])$} subject to the constraints resulting from $\alla^y[i]$ is equivalent solving the following problem:
\begin{alignat*}{2}
  &\Minimize_{\alphabase^{yy_{i}},\alphabase^{yy_{i+1}}} &&\sum_{a \in \cA} w_a \pi_a^y\max\{\max_{\hat{y} \in \cY_{i}}\eta(\alphabase^{y\hat{y}},\ta^{y\hat{y}}),\eta(\alphabase^{yy_{i+1}},\widetilde{\alpha}_a^{yy_{i+1}})\}\\
  &\st &&\alphabase^{yy_{i}},\alphabase^{yy_{i+1}}\geq 0 \text{ and } \alphabase^{yy_{i}}+\alphabase^{yy_{i+1}}=\gamma.
  \end{alignat*}
  Letting $v_a := \max_{\hat{y} \in \cY_{i-1}} \eta(\alphabase^{y\hat{y}},\alpha_a^{y\hat{y}})$, this is equivalent to
  \begin{align*}
    &\Minimize_{\alphabase^{yy_{i}}\in[0,\gamma]}\sum_{a \in \cA} w_a \pi_a^y\max\{v_a, \eta(\alphabase^{yy_{i}},\ta^{yy_{i}}), \eta(\gamma-\alphabase^{yy_{i}},\widetilde{\alpha}_a^{yy_{i+1}})\}.
  \end{align*}
  Similarly to the case of binary classification, this objective is one-dimensional, and concave in each of the intervals defined by the inflection points of the $\eta$ instances and the values for which any two of the expressions in the maximum are equal. Denote this set of points by $M_i^y$. Then the objective above is minimized by one of the values in the following set: $M_i^y \cup \{\alpha_a^{yy_{i}}\}_{a \in \cA} \cup \{\widetilde{\alpha}_a^{yy_{i+1}}\}_{a \in \cA} \cup \{0, \gamma\}$. Repeating this procedure until iteration \mbox{$i = k-1$}, we obtain an assignment for $\alla^y$ which can be used to calculate an upper bound for $\UNF_y(\allabigg^y)$.

  Since the ordering of the labels in the greedy procedure is arbitrary, it is possible to attempt several different orderings and select the one that obtains the smallest $\UNF$ value. In our experiments, we tried $10$ random orderings in each upper bound calculation.

\section{More Details on the Local Optimization Procedure} \label{app:OPT2}
Here we provide more details on the minimization of \eqref{obj_constraints_Large}. Given the split in \eqref{splitEta}, the last constraint in \eqref{obj_constraints_Large} becomes 
$$
\eta(\alpha_{\hat{y}}^y, h^y_{a,\hat{y}})\leq c_a^y \Longleftrightarrow \left\{
\begin{array}{c}
\eta_1(\alpha_{\hat{y}}^y, h^y_{a,\hat{y}})\leq c_a^y\\
\eta_2(\alpha_{\hat{y}}^y, h^y_{a,\hat{y}})\leq c_a^y
\end{array}
\right.,
$$
and now the constraints does not include singularity points, and can be locally approximated by two linear functions, one for $\eta_1$ and one for $\eta_2$ (see \figref{TaylorEta}). Explicitly, the Taylor approximation of $\eta_i$ for $i=1,2$ are given by:
$$
\eta_i(\alpha+\epsilon_a,b+\epsilon_b) \approx \eta_i(\alpha,b) + \frac{\partial\eta_i}{\partial \alpha}\epsilon_\alpha + \frac{\partial\eta_i}{\partial b}\epsilon_b,
$$
 where $\frac{\partial\eta_1}{\partial \alpha} = \frac{b}{\alpha^2}$,
$\frac{\partial\eta_1}{\partial b} =  - \frac{1}{\alpha}$,
$\frac{\partial\eta_2}{\partial \alpha} = \frac{-(1-b)}{(1-\alpha)^2}$, and $\frac{\partial\eta_2}{\partial b} = \frac{1}{1-a}$.

Given the first order Taylor approximations above, we form the LP approximation of \eqref{obj_constraints_Large} around an iterate $\bfx^{(t)} = \{\tilde \bfalpha^y,\tilde\bfH^y,\tilde\bfc^y\}_{y=1}^k$ as follows:
\begin{alignat}{2}\label{eq:obj_constraints_LP_Large}
&&&\Minimize_{
\{\bfalpha^y\},\{\bfH^y\},\{\bfc^y\}
}\quad
\sum_y \langle \bfw^y, \bfc^y\rangle \\
&&\st\,\,\,  &  0 \leq \bfalpha^y \leq 1 \quad\forall y\in[k], \nonumber \\
&&&  0 \leq \bfH^y \leq 1 \quad\forall y\in[k], \nonumber \\
&&&  0 \leq \bfc^y \leq 1 \quad\forall y\in[k], \nonumber \\
&&&   \langle \bfalpha^y, {\bf 1}_k\rangle=1 \quad\forall y\in[k], \nonumber\\
&&&   \bfH^y \mathbf{1}_k = \mathbf{1}_{|\mathcal{A}|} \quad\forall y\in[k],\nonumber\\
&&&   \textstyle \sum_{y=1}^k \pi^y_ah^y_{a,\hat{y}} = p_a^{\hat{y}} \quad\forall \hat{y}\in[k], a\in[|\mathcal{A}|]. \nonumber\\
&&&   \eta_i(\tilde\alpha_{\hat{y}}^y, \tilde h^y_{a,\hat{y}}) + \textstyle{\frac{\partial \eta_i}{\partial \alpha} (\alpha_{\hat{y}}^y-\tilde\alpha_{\hat{y}}^y)}   + \textstyle{\frac{\partial \eta_i}{\partial b} 
(h_{a,\hat{y}}^y-\tilde h_{a,\hat{y}}^y)}  \leq c_a^y,\quad\forall y,\hat{y}\in[k],  \;a\in[|\cA|], i=1,2,\nonumber
\end{alignat}

\eqref{obj_constraints_LP_Large} is an LP problem which is solved at each iteration by an LP solver, for which we use the {\tt scipy.optimize} library. In addition to the box constraints of $[0,1]$ for all variables, it includes $2\cdot k\cdot|\mathcal{A}| + k$ equality constraints, and $2\cdot k^2\cdot|\mathcal{A}|$ inequality constraints. It can be seen that the number of inequality constraints is rather high, and as a result, so is the computational complexity of the algorithm if run as is. However, many of these constraints are not active in the solution, and in any case we limit the step size of our algorithm. Hence, we can ease the difficulty of the LP problem by both limiting the search space of the LP solver, and removing the inequality constraints that seem to be inactive in the solution. To this end, we remove the inequality constraints where $\eta_i(\tilde\alpha_{\hat{y}}^y, \tilde h^y_{a,\hat{y}}) < -1$, as we expect these not to be active at the solution. Furthermore, we use a maximum step size $\tau$ so that
$\|\bfalpha^y - \tilde\bfalpha^y\|_{\infty} < \tau$, and $\|\bfH^y - \tilde\bfH^y\|_{\infty} < \tau$, 
where $\|\cdot\|_{\infty}$ is the maximum norm. This condition is trivially incorporated in the box constraints of \eqref{obj_constraints_LP_Large}. Finally, given the LP approximation above, the solver of \eqref{obj_constraints_Large} follows the same lines as Alg. \ref{alg:FairV1}.

As in \secref{multunknown}, numerical instabilities arise when $\alpha_a^{y\hat{y}}$ is too close to $0$ or $1$. Thus, here as well we restrict these values to be in the segment $[\varepsilon, 1-\varepsilon]$, this time as optimization variables through the box constraints. Also, we update the values $p_a^{\hat{y}}$ to be $p_a^{\hat{y}} = (1-k\epsilon)p_a^{\hat{y}} + \epsilon$ to guarantee a solution for \eqref{obj_constraints_Large} after updating the box constraints for $\alpha_a^{y\hat{y}}$.

\begin{figure}[h]
    \begin{center}
    \includegraphics[width=0.4\columnwidth]{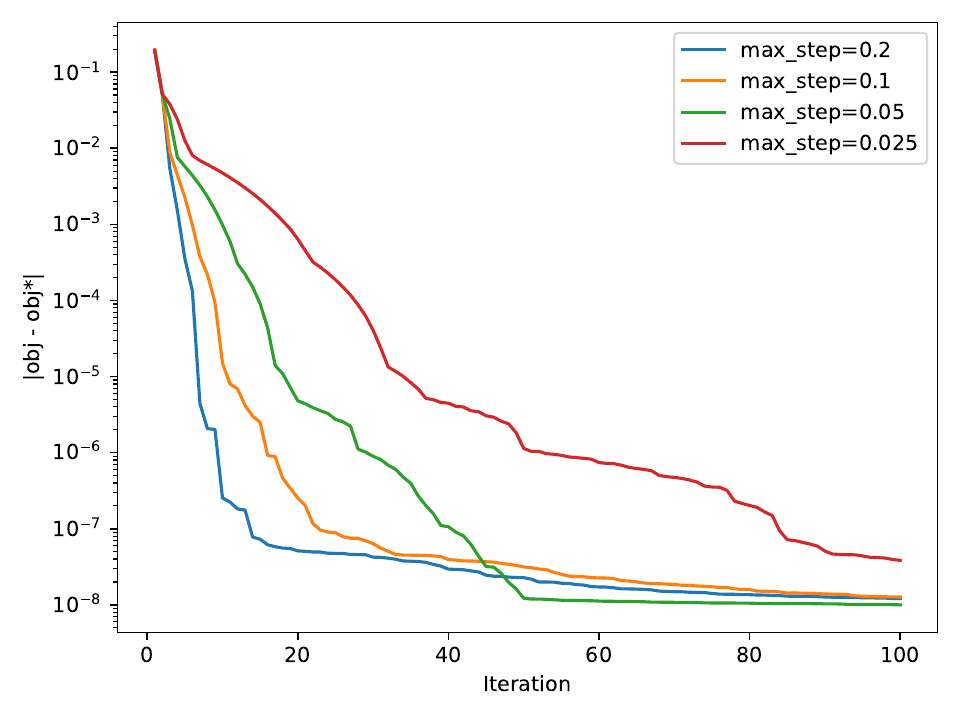}
    \caption{The convergence of Algorithm \ref{alg:FairV1} for different trust region (maximal step size) parameters $\tau$.}
  \label{fig:conv}
\end{center}
\end{figure}

\paragraph{Convergence of Algorithm \ref{alg:FairV1}} Our problem contains the non-linear functions $\eta$, which are approximated by linear functions. For a general non-linear programming problem, sequential linear programming (SLP) methods converge linearly if all functions are smooth. That is because they do not use Hessian information, and their whole purpose is to handle the constraints efficiently (determine who is active and who is not). If one uses Hessian information, we get sequential quadratic programming (SQP) and the convergence is asymptotically quadratic once the active set of constraints is identified, similar to Newton’s method \cite{nocedal2006numerical}. For an SLP method like in Algorithm \ref{alg:FairV1} to converge linearly, we first need to have smooth constraints, and for that we use the split of 
 presented in \eqref{splitEta} and detailed above. Furthermore, we also limit the size of the steps to be of size at most $\tau$. This approach is called a Trust Region method, and is commonly used with sequential quadratic or linear programming methods. The method was analyzed in \cite{kiessling2022feasible} for a case that is similar to ours, where the objective of the constrained problem is linear, like in \eqref{obj_constraints} and \eqref{obj_constraints_Large}. It is shown that the convergence of the SLP method is linear, and the rate depends on the radius of the trust region method ($\tau$, in our case). In Figure \ref{fig:conv} we show the convergence of our method for solving \eqref{obj_constraints_Large} for a convex $\eta$, so that the problem is convex and the solutions that are obtained by the method using all trust regions are equivalent. Specifically, to demonstrate the convergence we used 
 \begin{equation}\label{eq:etaconv}
\eta_{convex}(a, b) = \begin{cases}
a^2 - ab & b < a,\\ 
(1-a)^2-(1-b)(1-a) & b > a,\\
0 & b=a,
\end{cases}
\end{equation}
 instead of \eqref{etadef}, and got the plots in Figure \ref{fig:conv} for various maximal step sizes $\tau$, for solving \eqref{obj_constraints_Large} for the labor dataset. Here, the plots show that the method is slower as $\tau$ is smaller, since the algorithm is constrained to be slower. However, in some cases, the inner LP solution can result in a stagnate direction, and then decreasing $\tau$ as shown in \cite{kiessling2022feasible} can mitigate this. In our original problem, $\eta$ and the whole problem are non-convex, hence the algorithm can converge to a different local minimum for each value of $\tau$. For the problems reported in this paper, we used $\tau = 0.2$ which seemed to work best, and decreased $\tau$ when the LP solver failed to converge.

\end{document}